\documentclass{article}

\usepackage[preprint]{neurips_2021}

\newcommand{\piou}{\text{ProbIoU}}

\newcommand{\etal}{{et al}.\@ }
\newcommand{\red}[1]{{\color{red}#1}}
\newcommand{\blue}[1]{{\color{blue}#1}}

\usepackage{color,amsmath,amsfonts,comment}
\usepackage{multicol,subfig}
\usepackage{marvosym}
\usepackage{multirow}
\usepackage{times}
\usepackage{epsfig}
\usepackage{graphicx}
\usepackage{amsmath}
\usepackage{amssymb}
\usepackage{bm}

\usepackage[utf8]{inputenc} 
\usepackage[T1]{fontenc}    
\usepackage{hyperref}       
\usepackage{url}            
\usepackage{booktabs}       
\usepackage{amsfonts}       
\usepackage{nicefrac}       
\usepackage{microtype}      
\usepackage{xcolor}         

\title{Gaussian Bounding Boxes and Probabilistic Intersection-over-Union for Object Detection}

\author{%
  Jeffri M. Llerena \qquad Luis Felipe Zeni \qquad Lucas N. Kirsten \qquad Claudio Jung
  \\ \\
  Department of Informatics\\
  Federal University of Rio Grande do Sul\\
  Porto Alegre - RS, 91509-900, Brasil \\
  \footnotesize
  \texttt{\{jeffri.mllerena, luis.zeni, lnkirsten, crjung\}@inf.ufrgs.br}
}

\begin{document}

\maketitle

\begin{abstract}
  Most object detection methods use bounding boxes to encode and represent the object shape and location. In this work, we explore a fuzzy representation of object regions using Gaussian distributions, which provides an implicit binary representation as (potentially rotated) ellipses. We also present a similarity measure for the Gaussian distributions based on the Hellinger Distance, which can be viewed as a Probabilistic Intersection-over-Union (ProbIoU). Our experimental results show that the proposed Gaussian representations are closer to annotated segmentation masks in publicly available datasets, and that loss functions based on ProbIoU can be successfully used to regress the parameters of the Gaussian representation. Furthermore, we present a simple mapping scheme from traditional (or rotated) bounding boxes to Gaussian representations, allowing the proposed ProbIoU-based losses to be seamlessly integrated into any object detector.
\end{abstract}

\section{Introduction}

Object detection is a crucial problem in computer vision, and important advances have been made in the past years with the advance of deep learning methods~\cite{liu2020deep}. Most of these improvements are related to the use of larger or more challenging datasets (such as the Large Vocabulary Instance Segmentation -- LVIS dataset~\cite{gupta2019lvis}), dealing with class imbalance~\cite{oksuz2020imbalance}, proposing modern backbones such as multi-resolution feature extraction~\cite{lin2017feature,liu2018path,tan2020efficientdet}, and more recently the introduction of  transformers~\cite{carion2020end} or LambdaNetworks~\cite{bello:iclr:2021} for modelling longer-range interactions in the context of object detection and the analysis of the trade-off between classification and localization~\cite{wu2020rethinking}. However, the vast majority of object detection techniques explore horizontal (axis-aligned) bounding boxes (HBBs) to encode the object shape/location (because of its simplicity to implement and annotate training data), which are clearly not adequate when the objects do not present an aligned rectangular shape. Oriented Bounding Boxes (OBBs)~\cite{Chen:EECV:2020} can handle elongated objects with rotations, but the underlying representation is still rectangular.

As important as the strategy for representing the object location is the choice of a suitable function that can provide a distance or similarity between two objects both to evaluate detection results (test phase) and to regress the object representations (train phase). The Intersection over Union (IoU)~\cite{everingham2010pascal} has been adopted by the computer vision community as the most popular metric for comparing bounding boxes. In terms of HBB regression, most object detectors explore different norms ($\ell_1$, $\ell_2$, Huber) applied to either the HBB parameters or corners. More recently, some approaches have explored variations of the IoU itself as a loss function, such as the Generalized IoU (GIoU)~\cite{rezatofighi2019generalized} or the Distance Iou (DIoU)~\cite{zheng2020distance}, which aim to alleviate the problem of vanishing IoU gradients caused by disjoint HBBs or accelerate the convergence. However, they still assume a traditional HBB representation for the objects. Other approaches such as the Pixel-IoU (PIoU)~\cite{Chen:EECV:2020} can also handle oriented objects using OBBs, and recently Yang and colleagues proposed to map OBBs to Gaussian distributions and explored the Gaussian Wasserstein Distance (GWD)~\cite{yang2021rethinking} for the regression loss.

\begin{figure}[htb]
    \centering
    \includegraphics[height = 1.55cm]{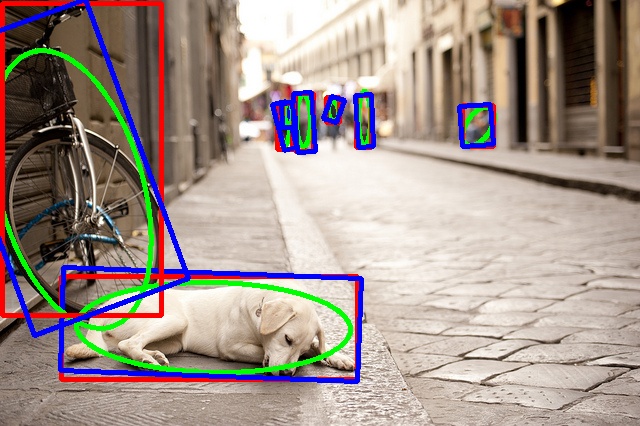}   \includegraphics[height = 1.55cm]{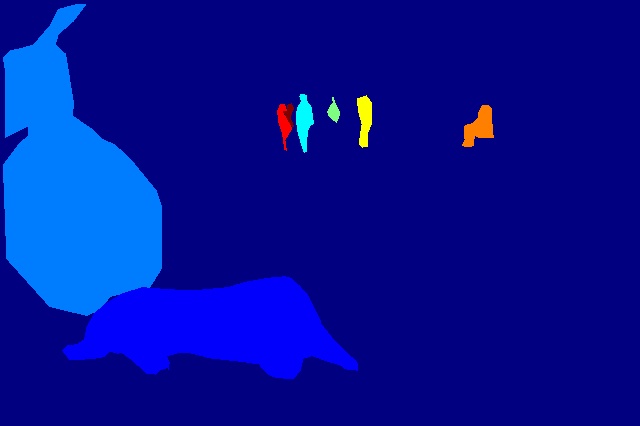}     \includegraphics[height = 1.55cm]{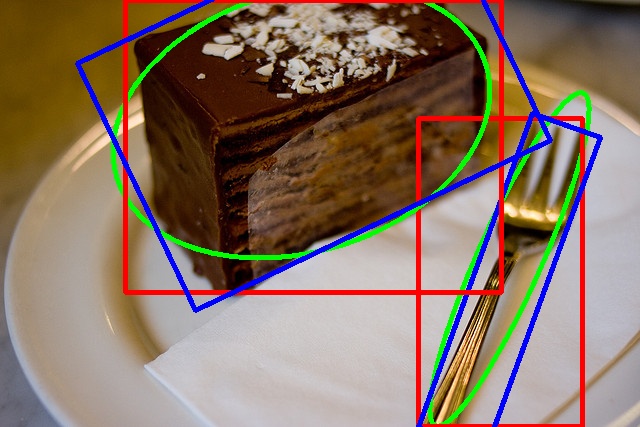}
    \includegraphics[height = 1.55cm]{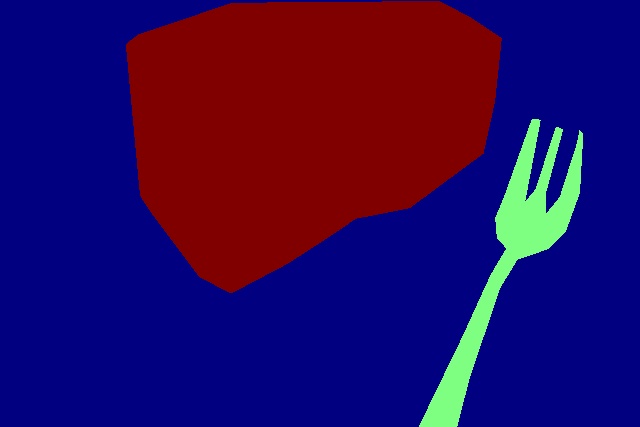}
    \includegraphics[height = 1.55cm]{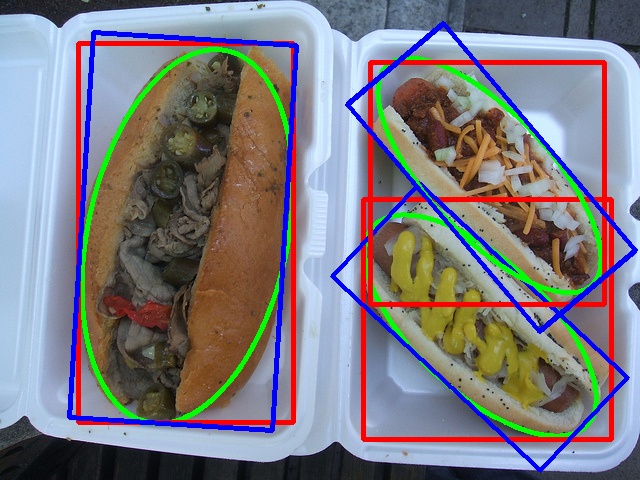}  \includegraphics[height = 1.55cm]{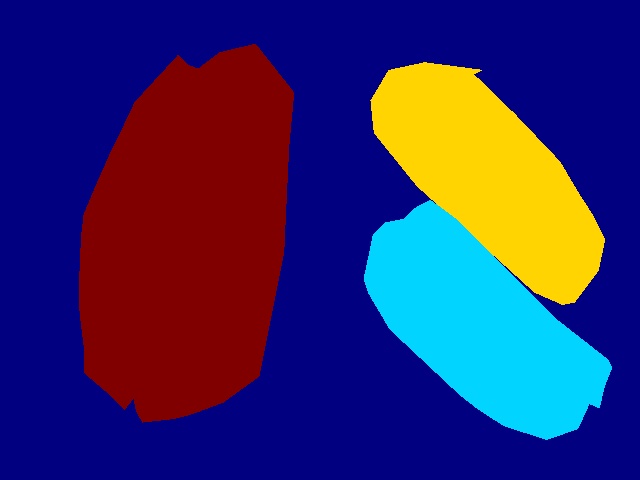}
    \caption{Different annotation types for images in the COCO dataset: HBBs (red), OBBs (blue) and  GBBs (green), as well as the GT segmentation masks.}
    \label{fig:problem:bb}
\end{figure}

In this work, we explore an object shape parametrization based on Gaussian Bounding Boxes (GBBs), and propose a ``Probabilistic IoU'' (called \piou{}) that emerges naturally from the use of Gaussian distributions to compare the similarity between objects. The proposed parametrization allows an intrinsic representation of the objects as (possibly rotated) ellipses, being more generic than traditional HBBs and even OBBs. As an example,  Figure~\ref{fig:problem:bb} shows some images from the COCO 2017 dataset~\cite{lin2014microsoft} with object location encoded as an HBB (red), OBB (blue) and the proposed GBB-induced ellipses (green), along with the corresponding segmentation masks. Note that OBBs do improve over HBBs for elongated and rotated objects (e.g., the fork in the middle image pair), but the proposed representation provides a tighter fit to the segmentation mask. As the main contribution we i) show that the GBB-induced elliptical representation tends to provide a tighter fit to segmentation masks than HBBs or OBBs; ii) introduce regression loss functions based on \piou{} that are intuitive, differentiable, simple to compute, and that can be seamlessly integrated into existing detectors that work with HBB or OBB.

\section{Related Work}


\textbf{Object representation:} HBBs are by far the most popular representation for objects, since it allows both simple representation  -- only four parameters -- and annotation cost, which is a concern when dealing with large datasets. On the other hand, objects can be represented as full binary masks, which provide a rich description of the objects, but requires tedious and time-consuming annotation. Furthermore, the networks should regress binary maps as well (as in semantic or panoptic segmentation tasks~\cite{kirillov2019panoptic}), resulting in more complex architectures. 
A well-known problem of HBBs is that they might cover considerable portions of the background (recall Figure~\ref{fig:problem:bb}), particularly for elongated and rotated objects. For this reason, Oriented Bounding Boxes (OBBs) have also been used as an alternative in between HBBs and segmentation masks. Ding~\etal~\cite{Ding:CVPR:2019} proposed a Rotated RoI Transformer to adapt a horizontal RoI to an OBB RoI; He and Chen~\cite{He:ICCV:2015, Chen:ICCV:2019} cover the problem of OBB proposals; Ma and colleagues~\cite{ma2019ship} explored OBBs for ship detection, while Chen~\etal~\cite{Chen:EECV:2020} proposed a new dataset with oriented objects (\verb'Retail50k') and explored OBBs with a novel Iou-based loss function for object detection. It is important to note that OBBs require an additional (angular) parameter to be regressed, and possibly a larger number of anchors (if an anchor-based detector is used). Moreover, regressing the angle parameter can be tricky since it depends on how the OBB rotation is defined, while loss functions based on the IoU cannot be used since they are not differentiable~\cite{yang2021rethinking}. Hence, defining a differentiable loss function that is agnostic to the rotation and relates to the IoU would be a valuable contribution.

\textbf{Localization Loss:} 
most object detectors based on HBBs apply a combination of parameter-wise loss components (e.g., using the $\ell_1$ or $\ell_2$ norms) to build the localization loss. Since these norms are highly sensitive to the scale of the objects, different strategies such as using the $\ell_2$ norm with HBB offsets in log space~\cite{Girshick:ICCV:2015}, or exploring  a smooth $\ell_1$ norm (also know as Huber loss)~\cite{Girshick:TPAMI:2015} have been proposed. Parameter-wise loss functions implicitly assume independence (i.e., changes in one parameter do not affect the others), so that the corners of an HBB might not converge consistently to the ground-truth (GT) annotation. As an alternative, some researchers have proposed IoU-based loss functions with well-known properties (e.g., invariance to scale changes and joint regression of the HBB parameters). UnitBox~\cite{Yu:ACM-ICM:2016} was one of the pioneers to incorporate an IoU loss to improve BB detection, while  Rahman~\etal~\cite{Rahman:IJVC:2016} used it for single object segmentation. Later, Tychsen-Smith and Petersson~\cite{Tychsen-Smith:CVPR:2018} proposed a bounding IoU loss for object detection, and Rezatofighi~\etal~\cite{rezatofighi2019generalized}
introduced the Generalized IoU (GIoU) to mitigate the problem of vanishing gradients produced by the IoU when the regions of the regressed and GT box do not overlap. Zheng~\etal~\cite{zheng2020distance} argued that GIoU presents slow convergence and proposed the Distance-IoU (DIoU) and the Complete-IoU (CIoU), which incorporate overlap area, distance between central points, and aspect ratio constraints as penalties to IoU. Sun~\etal~\cite{Sun:Access:2020} combined $\ell_n$ norms and IoU-based losses to tackle the drawbacks of each individual component, while Zhang~\etal~\cite{zhang2021focal} incorporated the  focal loss to the training and penalty to the IoU as the overlap area, the central point and the side length. 
Chen and colleagues~\cite{Chen:EECV:2020} presented an extension of the IoU to deal with Oriented Bounding boxes (OBBs), called Pixels-IoU (PIou), which allows a more generic object representation. While writing this paper, we became aware of the method proposed by Yang et al.~\cite{yang2021rethinking}, which models OBBs as Gaussian distributions (equivalent to our GBBs) and proposed a differentiable loss based on the Gaussian Wasserstein Distance (GWD). Their main goal was to deal with the orientation parameter in OBBs, and explored GBBs as intermediate representations. However, as shown in~\cite{yang2021rethinking}, an empirical non-linear mapping with an additional hyper-parameter must be applied to the GWD for using it as a localization loss function, and no geometric relation to the IoU was presented.

In this work, we explore Gaussian Bounding Boxes (GBBs) to encode the object representation, as done in~\cite{yang2021rethinking}. However, we do not assume any underlying representation as HBB or OBB to obtain the GBB, which leads to its application with generic segmentation masks as well. We also propose a similarity function that can be viewed as a probabilistic IoU (\piou{}) between any two distributions, and, for the particular case of Gaussians, it allows the definition of simple and differentiable loss functions. We also present a parametrization for the GBB representation that enables the adaptation of detectors that deal with OBBs (e.g., regressing center, dimension, and orientation, as in~\cite{yang2021rethinking}), or that regress directly the GBB parameters (mean vector and covariance matrix). 

\section{Gaussian Bounding Boxes and \piou{}}
\label{sec:gbbs}

The core idea in this paper is to represent an object in a fuzzy manner by using a two-dimensional, possibly rotated, Gaussian distribution. Such distributions are characterized by the mean vector $\bm{\mu} = (x_0, y_0)^T$ and the covariance matrix $\Sigma$, and they induce elliptical representations of the desired 2D regions.  By representing regions as probability distributions, we can explore a vast arsenal of statistical tools to compare the distance or similarity between two distributions. The particular case of Gaussian distributions leads to closed-form differentiable expressions for some of these metrics, such as the Bhatacharyya distance~\cite{bhattacharyya1943measure} and the Kullback-Leibler divergence~\cite{kullback1951information}, which are attractive options in deep learning strategies.

\subsection{Gaussian Bounding Boxes (GBBs)} \label{sec:gaussian_bounding_boxes}
Before providing details on how to represent 2D regions as GBBs, we discuss some properties of the covariance matrix and how the parameters can be regressed through a deep neural network. We first note that the mean vector $\bm{\mu} = (x_0, y_0)^T$ of a 2D Gaussian does not impose any restriction on the parameters $x_0, y_0$. 

A 2D covariance matrix $\Sigma$ presents three degrees of freedom, which can be expressed directly as the elements $a,b,c$ of the (symmetric) covariance matrix, or as the decorrelated variances $a',b'$ and the rotation angle $\theta$, which are related through
\begin{equation}
\label{cov:rotation}
    \Sigma = \begin{bmatrix}
    a & c\\
    c & b 
    \end{bmatrix} =  R_{\theta}\begin{bmatrix}
    a' & 0\\
    0 & b' 
    \end{bmatrix}R_{\theta}^T = \begin{bmatrix}a' \cos^{2}\theta + b' \sin^{2}\theta & \frac{1}{2}\left(a' - b'\right) \sin2 \theta \\
    \frac{1}{2}\left(a' - b'\right) \sin2 \theta  & a' \sin^{2}\theta  + b' \cos^{2}\theta\end{bmatrix},
\end{equation}
where $R_\theta$ is a two-dimensional rotation matrix. Let us recall that $\Sigma$ must be positive definite,  and Sylvester's criterion~\cite{gilbert1991positive} states that a Hermitian  matrix is positive-definite if and only if all the leading principal minors are positive. Such criterion implies that $a>0$ and $\det \Sigma = ab - c^2 >0$ (which also implies that $b > 0$), providing a set of restrictions if we want the network to regress $(a,b,c)$ directly. 

We can also design a network that regresses $(a',b',\theta)$ instead, which is more aligned to the output of existing object detectors for OBBs and allows an easier adaptation for GBBs, as noted in~\cite{yang2021rethinking}. In this parametrization, 
we must have $a',b'> 0$ and $\theta \in \mathbb{R}$, which leads to an ambiguous representation of the covariance matrix:  parameters $(a',b',\theta)$ and $(b', a', \theta + \pi/2)$ generate exactly the same matrix. To mitigate this problem, we can select $\theta \in [-\pi/4, \pi/4]$ to generate a unique representation of a covariance matrix $\Sigma$ in terms of $a',b',\theta$, which also solves the circularity problem of the angular parameter (i.e., $\theta$ and $\theta + 2\pi$ producing the same matrix). It is also important to note that we have an isotropic Gaussian when $a'=b'$, and in this case, the orientation $\theta$ becomes irrelevant (any value for $\theta$ yields the same matrix $\Sigma$). A square-shaped OBB, on the other hand, still provides an associated orientation, so that the mapping from a square-shaped OBB to a GBB is not an invertible operation.

\subsection{Converting HBBs and OBBs to the GBB representation}
Since most existing datasets for object detection present annotations as HBBs, OBBs or even segmentation masks, we present next some possibilities for obtaining GBBs from these representations. We start by assuming that the object region is represented as a 2D (continuous) binary region $\Omega$, which is the most complete shape description. Assuming that $\Omega$ represents a uniform Probability Density Function (PDF), the mean of the distribution and the covariance matrix are given by 
\begin{equation}
    \label{eq:mu}
    \bm{\mu} = \frac{1}{N}\int_{\bm{x}\in\Omega} \bm{x},~~
    \Sigma = \frac{1}{N}\int_{\bm{x}\in\Omega} \left(\bm{x} - \bm{\mu} \right)\left(\bm{x} - \bm{\mu} \right)^T,
\end{equation}
where $N = \#\Omega$ is the area of $\Omega$ (and hence $1/N$ is the density for each point in $\Omega$). For HBBs, $\Omega$ reduces to a rectangular region with center $(x_0,y_0)$, width $W$ and height $H$. In this case, $\bm{\mu} =(x_0,y_0)^T$ is simply the center of the rectangular region, and the covariance matrix can be computed analytically (after subtracting each point in $\Omega$ from the mean) through
\begin{equation}
\label{eq:cov:gbb}
\Sigma = \frac{1}{WH}\int_{-H/2}^{H/2}\int_{-W/2}^{W/2}  
\begin{bmatrix}
x^2 & xy \\
xy & y^2
\end{bmatrix}dxdy
= \frac{1}{12}\begin{bmatrix}
W^2 & 0 \\
0 & H^2
\end{bmatrix},
\end{equation}
so that $a = W^2/12,b = H^2/12$ and
$c=0$ (the inverse transformation is  straightforward).

If we have an OBB instead, we can define uncorrelated variances $a'$ and $b'$ based on the sides of the corresponding axis-aligned HBB as above, and also compute the orientation angle $\theta$. The corresponding covariance matrix is obtained using Eq.~\eqref{cov:rotation}.  If the annotation is provided as parametrized 2D shape (e.g., a polygonal representation), we might be able to compute the Gaussian parameters analytically using Eq.~\eqref{eq:mu}. For a generic representation using binary segmentation masks, we estimate the empirical mean and covariance matrix, which are essentially discretized versions of Eq.~\eqref{eq:mu}.

It is also important to provide strategies for generating binary masks from the fuzzy representations provided by the GBBs for visualizing the regions as deterministic masks and also for computing the traditional IoU between two regions. If the application requires an HBB or OBB representation, the box dimensions (and angle, if applicable) can be extracted from the diagonalized covariance matrix using the inverse of Eq.~\eqref{eq:cov:gbb}. We also present an alternative representation based on elliptical regions, as explained next.  

For a multivariate Gaussian distribution, the  Mahalanobis squared distance     $d^2(\bm{x}) = \left(\bm{x} - \bm{\mu} \right)^T\Sigma^{-1}\left(\bm{x} - \bm{\mu} \right)$ presents a chi-squared distribution (with two degrees of freedom in our case). Also, the level sets given by $d^2(\bm{x}) = r^2$ (with $r>0$) are represented as elliptical regions, for which the principal directions are the eigenvectors of $\Sigma^{-1}$, and the semi-axes are $r/\sqrt{\lambda_1}$ and $r/\sqrt{\lambda_2}$, where $\lambda_1,\lambda_2$ are the corresponding eigenvalues. Hence, each value of $r$ relates to a different ellipse, and one possibility for choosing $r$ is based on the area of the Gaussian that lies in the interior of the ellipse. Given an area threshold $0<\tau < 1$, we can set $r$ such that $F_{\chi^2}(r^2) = \tau$, where $F_{\chi^2}$ is the Cumulative Distribution Function (CDF) of the chi-squared distribution. If the GBB was computed from an HBB (or OBB) using Eq.~\eqref{eq:cov:gbb}, we can show that $r=\sqrt{12/\pi}$ yields an ellipse with the same area, and it corresponds to $\tau \approx 0.85$. This is our default mapping using in all experiments (and also used to generate the ellipses in Figure~\ref{fig:problem:bb}).

\subsection{The Probabilistic IoU (\piou{}) and Localization Loss Functions}
\label{sec:piou:locloss}

Given two objects (GT annotation and prediction) represented as GBBs, the next step is to define how to compute the similarity (or dissimilarity) between them. For that purpose, we can choose from several statistical tools devised for computing the overlap (similarity) or distance (dissimilarity) between two probability distributions~\cite{vegelius1986measures}.
In this work, we focus on the Bhattacharyya Coefficient/Distance, which can be used to obtain an actual distance metric, as discussed next.

The Bhattacharyya Coefficient $B_C$ between  two 2D probability density functions $p(\bm{x})$ and $q(\bm{x})$ is defined as
\begin{equation}
\label{eq:bc}
    B_C(p,q) = \int_{\mathbb{R}^2}\sqrt{p(\bm{x})q(\bm{x})}d\bm{x},
\end{equation}
and it measures the amount of \textit{overlap} between the distributions (hence, computing the \textit{similarity} between them). It is easy to see that $B_C(p,q) \in[0,1]$, and $B_C(p,q) = 1$ if and only if the two distributions are the same (when they are piece-wise continuous functions, which is the case of Gaussians). The Bhattacharyya Distance $B_D$ between two distributions $p$ and $q$ is given as 
\begin{equation}
    B_D(p,q) = -\ln B_C(p,q),
\end{equation}
and it provides an estimate of the \textit{dissimilarity} between them: when $B_D$ increases $B_C$ decreases, and vice-versa. When both $p\sim\mathcal{N}(\bm{\mu}_1, \Sigma_1)$ and $q\sim\mathcal{N}(\bm{\mu}_2, \Sigma_2)$ are Gaussian distributions with 
\begin{equation}
    \bm{\mu}_1 = \begin{pmatrix}
    x_1 \\ y_1
    \end{pmatrix},~
    \Sigma_1 = \begin{bmatrix}
    a_1 & c_1\\
    c_1 & b_1 
    \end{bmatrix},~
    \bm{\mu}_2  = \begin{pmatrix}
    x_2 \\ y_2
    \end{pmatrix},~
    \Sigma_2 = \begin{bmatrix}
    a_2 & c_2\\
    c_2 & b_2 
    \end{bmatrix},
\end{equation}
we can obtain a closed-form expression for $B_D$ given by
\begin{equation}
\label{eq:bhatta}
B_D = \frac{1}{8}(\bm{\mu}_1 - \bm{\mu}_2)^T\Sigma^{-1}(\bm{\mu}_1 - \bm{\mu}_2) +   \frac{1}{2}\ln \left( \frac{\det\Sigma}{\sqrt{\det\Sigma_1 \det\Sigma_2}}  \right), ~~\Sigma = \frac{1}{2}\left(\Sigma_1 + \Sigma_2\right).
\end{equation}

Since we are dealing with 2D vectors and matrices, the inverse and determinant can be computed analytically. More precisely, we can
write $B_D=B_1+B_2$ with
\begin{equation}
B_1 = \frac{1}{4}\frac{(a_{1} + a_{2}) (y_{1} - y_{2})^2 +(b_{1} + b_{2}) (x_{1} - x_{2})^{2} +2(c_1 + c_2)(x_2 - x_1)(y_1 - y_2)}{(a_1 + a_2)(b_1 + b_2) - (c_1 + c_2)^2}, 
\end{equation}
\begin{equation}
    B_2 = \frac{1}{2} \ln\left(\frac{(a_1 + a_2)(b_1 + b_2) - (c_1 + c_2)^2}{4 \sqrt{(a_1b_1-c_1^2)(a_2b_2-c_2^2)}} \right),
\end{equation}
and consequently we can also obtain an analytical closed-form expression for $B_C = e^{-B_D}$. Note that $B_2$ involves only the shape parameters $\Sigma_1$, $\Sigma_2$ and does not depend on the mean $\bm{\mu}_1$, $\bm{\mu}_2$,
so that alternative formulations that might prioritize central point adherence or shape consistency can be designed by weighing $B_1$ and $B_2$ differently.

It is important to mention that the Bhattacharyya Distance is not an actual distance since it does not satisfy the triangle inequality. However, the Hellinger distance defined as
\begin{equation}
   H_D(p,q) =\sqrt{1- B_c(p,q)}
\end{equation}
satisfies all the requirements for a distance measure~\cite{kailath1967divergence}, and can also be expressed analytically as a function of the Gaussian parameters. It is easy to see that $0 \leq H_D(p,q) \leq 1$, and for Gaussian distributions we always have $H_D(p,q) > 0$ when $p \neq q$ (since $B_C(p,q)$ is strictly positive due to the infinite support of the Gaussian).  In this work, we propose to use $1 - H_D(p,q)$ as a similarity measure between two Gaussian distributions, which can be viewed as a \textit{Probabilistic} IoU (or \piou{}) between the corresponding GBBs (the relation between \piou{} and IoU is further discussed in the supplementary material).  Note that the Gaussian Wasserstein Distance (GWD) used in~\cite{yang2021rethinking} does not present a clear relationship with the IoU, and an empirical mapping function must be used to explore the GWD as a localization loss.

Although  GBBs and \piou{} can be used in the future as an alternative way to validate object detection algorithms, in this work we are interested in using them for training a detector. Let us consider that $\bm{p} = (x_1,y_1,a_1,b_1,c_1)$ is the set of GBB parameters regressed by a network, and that $\bm{q} = (x_2,y_2,a_2,b_2,c_2)$ are the ground truth (GT) annotations for the desired GBB. We propose the following functions based on \piou{} to be used as the localization loss in an object detector:
\vskip .2cm
\noindent 
1.  $\mathcal{L}_1(\bm{p},\bm{q}) = H_D(\bm{p},\bm{q}) = 1 - \piou(\bm{p},\bm{q}) \in [0,1]$,
\vskip .1cm
\noindent 
2. $\mathcal{L}_2(\bm{p},\bm{q}) =  B_D(\bm{p},\bm{q}) = -\ln\left(1 - \mathcal{L}_1^2(\bm{p},\bm{q}) \right)\in [0,\infty]$,
\vskip .1cm
\noindent 
which are both differentiable with respect to $\bm{p}$ (and the gradient can be computed analytically), and reach the ideal minimum value zero iff $\bm{p}=\bm{q}$. 
Note that $\mathcal{L}_1$ follows the direction of~\cite{rezatofighi2019generalized,zheng2020distance,wang2020inferring}, who explore variations of the IoU for training object detectors. However, $\mathcal{L}_1$ might produce values very close to one when the GBBs are far from each other, which might lead to very small gradients and lead to slow convergence. On the other hand, $\mathcal{L}_2$ does not present this limitation, but it is not geometrically related to an IoU measure. Hence, we suggest to start training with $\mathcal{L}_2$ and then switch to $\mathcal{L}_1$.


\textbf{Axis-aligned GBBs:} if we want to deal with HBBs only, the corresponding axis-aligned GBBs would present diagonal covariance matrices (i.e., $c_1 = c_2 = 0$). This leads to only four parameters to be regressed (as in HBBs), and the only constraints are $a>0$ and $b>0$ to ensure positive-definiteness of the covariance matrix. If we have any object detector that produces HBB parameters (center point, width, and height),  we propose a very simple adaptation to use localization losses based on \piou{}. 

If $\bm{p}_{B} = (x_1,y_1,W_1,H_1)$ encodes the center $(x_1,y_1)$, width $W_1$ and height $H_1$ of an HBB produced by the object detector, then the corresponding GBB parameters are given by $\bm{p}_{G} =  (x_1,y_1,a_1,b_1) =  \bm{f}(\bm{p}_{B}) = (x_1,y_1,W_1^2/12,H_1^2/12)$, as proposed in  Section~\ref{sec:gaussian_bounding_boxes} -- note that $\bm{f}$ is  differentiable w.r.t. to the HBB parameters. A similar mapping $\bm{q}_{G} = (x_2,y_2,a_2,b_2) = \bm{f}(\bm{q}_{B})$ can be applied to the HBB parameters of a GT annotation, and we can apply a GBB-based localization loss $\mathcal{L}_{G}(\bm{p}_{G},\bm{q}_{G})$ in the transformed coordinates (such as $\mathcal{L}_1$ or $\mathcal{L}_2$  defined before). Note that both loss functions involve the term $B_D$, which provides a particularly simple expression for axis-aligned BGGs: 
\begin{equation}
B_D(\bm{p}_G, \bm{q}_G) = \frac{1}{4}\left(\frac{(x_1-x_2)^2}{a_1+a_2} +\frac{(y_1-y_2)^2}{b_1+b_2} \right) 
+\frac{1}{2}\ln\left( (a_1+a_2)(b_1+ b_2) \right) 
-
\frac{1}{4}\ln\left(a_1a_2b_1b_2\right), 
\label{eq:bhatta:diagonal}
\end{equation}
noting that the constant term  $ \ln 2$ was discarded since it does not affect the gradient. For the sake of comparison, the GWD loss proposed in~\cite{yang2021rethinking} reduces to a weighted parameter-wise $\ell_2$ loss when applied to axis-aligned GBBs (which is counter intuitive to its relation with IoU), whereas our loss jointly explores all the parameters.

For training the network using backpropagation, the gradient $\nabla \mathcal{L}_{G}$ is required, and it depends on both the Jacobian of $\bm{f}$ (which is diagonal and very easy to obtain) and the partial derivatives $\partial B_D/\partial \bm{p}_G$ of the Bhattacharyya Distance (i.e., $\mathcal{L}_2$), which are also straightforward. It can be shown that  $\partial B_D/\partial \bm{p}_G = \bm{0} \Leftrightarrow \bm{p}_G = \bm{q}_G \Leftrightarrow \bm{p}_B = \bm{q}_B$, which means that we cannot have vanishing gradients during training when using $\mathcal{L}_2$ as the localization loss. By using the chain rule, we can observe that $\mathcal{L}_1 = \sqrt{1 - \text{exp}(-B_D)}$ also does not present vanishing gradients, since $\partial\mathcal{L}_1/\partial B_D$ is always strictly positive. However, the exponential mapping used in $\mathcal{L}_1$ might generate numerically vanishing gradients, particularly in the beginning of training where regressed GBBs might be very far from the annotations. This observation corroborates our proposal of using a two-stage training protocol that starts with $\mathcal{L}_2$ for an initial alignment, and then switches to $\mathcal{L}_1$ for a better fit of the GBBs.


\vspace{.4cm}
\noindent
\textbf{Properties:}  the proposed \piou{} based on the Hellinger distance (and the associated loss functions) presents several properties that make it attractive as the regression loss in object detectors:
\begin{itemize}
    \item All the three functions are differentiable with respect to all parameters, as shown before;
    \item The Helinger distance satisfy all distance metric criteria~\cite{kailath1967divergence}, and hence so does $1-$ \piou{}.
    \item The loss functions are invariant to object scaling. If two object regions $\Omega_1$ and $\Omega_2$ that induce the distributions $p\sim\mathcal{N}(\bm{\mu}_1, \Sigma_1)$ and $q\sim\mathcal{N}(\bm{\mu}_2, \Sigma_2)$ are scaled according to a factor $s$, the corresponding distributions are $p'\sim\mathcal{N}(s\bm{\mu}_1, s^2\Sigma_1)$ and $q'\sim\mathcal{N}(s\bm{\mu}_2, s^2\Sigma_2)$. Then, it is easy to see that $B_D(p,q) = B_D(p', q')$ according to Eq.~\eqref{eq:bhatta}. Hence, the Bhatacharyya Coefficient, the Helinger Distance and  \piou{} are also invariant to rescaling.
\end{itemize}

\section{Experimental Results}
\label{sec:experimental}

We start this section by showing that GBB-induced elliptical masks are closer to GT segmentation masks than HBB or OBBs, and might be used in the future as an alternative to these traditional representations. Next, we show results using the proposed loss functions for object detectors using HBBs and OBBs, along with comparisons with SOTA approaches. Due to hardware limitations (more details on the used hardware in supplementary material), we could not always use the exact same training setup suggested in the papers/repositories (so that our results might not match the results reported in the original papers).  However, we used the same setup for each detector/dataset and changed only the localization loss, which allows a fair comparison of the individual effect of the localization loss in the final accuracy.

For the experiments with our loss functions, we defined a default setup where we start training with $\mathcal{L}_2$ loss for half of the total iterations, and then switch to $\mathcal{L}_1$. The only hyper-parameters of our method are the weights $\omega_1$
or $\omega_2$ related to $\mathcal{L}_1$ and $\mathcal{L}_2$, respectively, when combining them with the classification loss (as in most object detectors). Although we have not performed exhaustive experiments, we noted that choosing $\omega_2 \approx 5\omega_1$ maintains the gradient of the localization loss with similar magnitudes when switching from $\mathcal{L}_2$ to $\mathcal{L}_1$. When we define a single weight $\omega$ for a given test, we mean that $\omega_1=\omega$ and $\omega_2=5\omega$.
%

\subsection{Comparing different object representations}
We evaluated all images in the training set from COCO 2017, and considered that the provided segmentation masks are the ground truth in terms of object localization
(masks with more than one connected component related to a single object, mostly due to occlusions, were excluded in our analysis). From the segmentation masks we computed the HBBs, the OBBs (following~\cite{Chen:EECV:2020}) and also the proposed elliptical GBB representations, and evaluated the IoU between each representation and the GT masks for each category present in the dataset. Our results indicate that using elliptical regions yields higher median IoU values than HBBs and OBBs for the vast majority  of the categories (77 out of 80). The exceptions were \verb'traffic light' and \verb'microwave', for which the HBB representation was the best, and \verb'tv', for which OBB was the best. Moreover, we found out that in 31 out of 80 categories, the median IoU values using HBBs were smaller than 0.5, which means that considerable portions of the background are indeed present in the HBB representation. This means that even if an HBB-based detector is able to regress the GT annotation perfectly, it would still fall below the 0.5 IoU threshold when compared to the segmentation mask (i.e., it would be considered a false positive). For OBBs, the number of categories in this situation is 13, and for GBBs, only one: \verb'kite'. More details on this experiment are provided in the supplementary material.

\subsection{Results for object detection using HBBs}

To show that the proposed \piou{}-based loss functions can be used as an alternative for object detectors based on HBBs, we performed a comparison
using two popular popular detectors (EfficientDet~\cite{tan2020efficientdet} and SSD~\cite{Liu:ECCV:2016}) trained with our loss, the usual smooth $\ell_1$ loss, and IoU-based losses, namely GIoU~\cite{rezatofighi2019generalized}, DIoU and CIoU~\cite{zheng2020distance}. In all experiments, we used the PASCAL VOC 2007 dataset~\cite{everingham2010pascal}, which was also evaluated in~\cite{rezatofighi2019generalized,zheng2020distance}.
In the experiments with EfficientDet, we used a Tensorflow 2 implementation\footnote{https://github.com/xuannianz/EfficientDet} with the D0 backbone; for SSD, we adopted a PyTorch implementation\footnote{https://github.com/Zzh-tju/DIoU-SSD-pytorch} with ResNet-50-FPN as backbone and $300\times 300$ image resolution. Both models were trained and evaluated in the same VOC2007 testset.

The training setup for the experiments are described below. All parameters were inherited from the used implementations (repository), unless otherwise stated.
\begin{itemize}
    \item \textbf{EfficientDet D0:} We used the Adam optimizer~\cite{Kingma:ICLR:2014} ($\epsilon=10^{-3}$, $\beta_1=0.9$, $\beta_2=0.999$) with initial learning rate$=10^{-3}$ and using a clip value of $10$ for the gradients; batch size$=12$; and the focal loss ($\alpha=0.25$, $\gamma=1.5$) for the classification loss. The model was trained for 50 epochs with the backbone layers frozen (i.e. no gradient propagation), and then for more 50 epoch training all layers. We added to our implementation a \emph{plateu learning rate reducer factor} of $0.8$ for the validation loss, with 2 epochs patience; and the training weights were saved every epoch the model improved in the validation loss. This showed to return better results for all tested losses.
    
    \item \textbf{SSD300:} We used Stochastic Gradient Descent with learning rate$=10^{-3}$, momentum$=0.9$, weight decay$=5\times 10^{-4}$ and using a clip value of $10$ for the gradients; batch size$=32$ for 220 epochs. For classification we use focal loss ($\alpha=1$, $\gamma=2$),  weight of $5$ for IoU-based regression loss functions (as done in~\cite{zheng2020distance}), and $\omega = 2$ for our approach.  The training weights were saved every ten epochs or whenever the validation loss improved. 
\end{itemize}

We evaluated the models using the AP75 and AP (i.e., AP$_{50:95}$) metrics with both the traditional IoU and our proposed \piou{} (i.e., $1-\mathcal{L}_1$) as evaluation metrics, as shown in Table~\ref{table:hbb_results} (in all tables, the best value is shown in red, and the second best in blue). As expected, our method yielded the best values using \piou{} as the evaluation metric (as expected, since it is explicitly used as the localization loss), and also competitive results when using the traditional IoU-based metrics.

\begin{table}[ht!]
    \centering
    \footnotesize
    \caption{Results HBB object detectors in PASCAL-VOC 2007 dataset}
    \begin{tabular}{c|cc|cc|cc|ccc}
        & \multicolumn{4}{c|}{EfficientDet D0}  & \multicolumn{4}{c}{SSD300} \\ \cline{2-9}
        \multirow{2}{*}{Loss} & \multicolumn{2}{|c}{AP} &  \multicolumn{2}{c|}{AP75}  & \multicolumn{2}{|c}{AP} &  \multicolumn{2}{c}{AP75}  \\  \cline{2-9}
        & \textbf{IoU}  & \textbf{ProbIoU} & \textbf{IoU} & \textbf{ProbIoU} & \textbf{IoU}  & \textbf{ProbIoU} & \textbf{IoU} & \textbf{ProbIoU}      \\ \hline
        \textbf{Smooth$\,\ell_1$}  & 40.72 & 54.49 & 42.02 & 61.26 & 40.53 & 60.07 & 42.75  & 67.24 \\  
        \textbf{GIoU} & 42.23 & 55.33 & 43.96 & 62.12  & \red{42.68} & \blue{64.15} & \blue{43.72}  & \blue{72.34} \\  
        \textbf{DIoU} & \blue{42.64} & 55.31 & \blue{44.74}  & 62.26  & 41.99 & 63.95 & 42.96  & 72.27 \\  
        \textbf{CIoU} & \red{42.94} & \blue{55.87} & \red{45.35} & \blue{63.04}  & \blue{42.50} & 64.07 & \red{44.80} &  72.07 \\  
        \textbf{Ours$_{\omega=2}$} & 42.60 & \red{56.76} & 44.24 & \red{64.15} & 41.89 & \red{64.16} & 43.00 & \red{72.39}
    \end{tabular}
    \label{table:hbb_results}
\end{table}

\subsection{Results for object detection using OBBs}

For the oriented object detection scenario, we explored two popular datasets that contain OBB annotations for the objects: 
\begin{itemize}
    \item \textbf{DOTA}~\cite{Xia:2018:CVPR} is comprised of 2,806 large aerial images from different sensors and platforms. Objects in DOTA exhibit a wide variety of scales, orientations, and shapes, being     annotated by experts into 15 object categories. The short names for categories are defined as (abbreviation-full name): PL-Plane, BD-Baseball diamond, BR-Bridge, GTF-Ground field track, SV-Small vehicle, LV-Large vehicle, SH-Ship, TC-Tennis court, BCBasketball court, ST-Storage tank, SBF-Soccer-ball field, RA-Roundabout, HA-Harbor, SP-Swimming pool, and HCHelicopter. We used the version 1 (v1) of this dataset, and we used the same procedures described in~\cite{yang2021rethinking}, in order to employ fair comparisons.
    \item \textbf{HRSC2016}~\cite{liu:icpram:2017} contains images from two scenarios including ships on sea and ships close inshore. The training,
validation and test set include 436, 181 and 444 images, respectively.
\end{itemize}
For each dataset, we compared two SOTA OBB object detectors: RetinaNet~\cite{lin2017focal} and R$^3$~Det~\cite{yang2021r3det} with a ResNet50~(R-50) backbone and the addition of an angle regression component in both detector heads. For each detector/dataset, we evaluated the results using the proposed loss functions and the GWD-based loss recently proposed in~\cite{yang2021rethinking}, which showed better results than the smooth $\ell_1$ loss. We used the implementation of both OBB detectors from the UranusDet Benchmark\footnote{https://github.com/yangxue0827/RotationDetection}~\cite{yang2021rethinking}.

Since our hardware does not support the default training setup proposed in~\cite{yang2021rethinking} (we had to reduce the batch size), we re-trained the models with their GWD loss function using the setup used on our loss. Moreover, in order to avoid exhaustively long training periods (and also more CO$_2$ emissions), we also reduced the total number of iterations. The evaluation protocol was based on the guidelines of each dataset. 

For DOTA v1, we analyzed the per-class AP50 accuracy as well as the average AP50 (i.e. the mean AP50 value considering all classes), as shown in Table~\ref{table:dota_gbb_results}. GWD-ret relates to our re-trained version of GWD, whereas GWD-rep indicates the values reported in~\cite{yang2021rethinking}. Using the R-50 Retinanet detector, our AP was more than 2\% higher than GWD-ret, and very close to GWD-rep (being better in 6 individual classes). In R$^3$ Det, our AP was below than GWD-rep and GWD-ret, but it is still competitive (being better in 2 individual classes).

For the HRSC2016 dataset, we evaluated the global AP (using different IoU acceptance thresholds) as shown in Table~\ref{table:hrsc_gbb_results}. Although the batch size (BS) had to be significantly reduced (to a single sample), the results with our combination of $\mathcal{L}_1$ and $\mathcal{L}_2$ produced results comparable to the GWD-rep (which was expored to $12\times$ more total samples than our train), and better than the re-trained version in the AP metrics computed with different IoU thresholds.  A similar behavior can be observed when using R$^3$~Det, where the proposed loss yields an AP50-90 value more than 5\% higher than GWD (in the re-trained experiment).


\addtolength{\tabcolsep}{-4pt} 
\begin{table*}[ht!]
    \centering
    \tiny
    \caption{Results for OBB detection in the DOTA v1 dataset}
    \begin{tabular}{c|l|ccccccccccccccc|c|rr}
        \\
        Model & Loss & PL & BD & BR & GTF & SV & LV & SH & TC & BC & ST & SBF & RA & HA & SP & HC & AP50 & Iterations & BS\\ \hline
        \multirow{3}{*}{R-50 Retinanet}
        & GWD-rep & 88.49 & \red{77.88} & \red{44.07} & \blue{66.08} & \red{71.92} & \blue{62.56} & \blue{77.94} & \blue{89.75} & \red{81.43} & \blue{79.64} & \red{52.30} & \red{63.52} & \blue{60.25} & \red{66.51} & \blue{51.63} & \red{68.90} & 1,080,000 & 8 \\
        
        & GWD-ret & \blue{88.77} & 71.76 & \blue{42.25} & 60.05 & 68.35 & 59.04 & 75.06 & \red{89.90} & 77.26 & 78.22 & 48.93 & \blue{63.11} & 54.86 & 64.09 & \red{51.81} & 66.23 & 351,000 & 2 \\
        
        & Ours$_{\omega=2}$ & \red{88.80} & \blue{71.97} & 41.90 & \red{68.45} & \blue{71.06} & \red{69.22} & \red{79.75} & 89.61 & \blue{78.84} & \red{80.15} & \blue{50.79} & 61.96 & \red{60.84} & \blue{64.59} & 49.07 & \blue{68.47} & 351,000 & 2
        \\ \hline
        \multirow{3}{*}{R-50 R$^{3}$Det}
        & GWD-rep & \blue{88.82} & \red{82.94} & \red{55.63} & \red{72.75} & \red{78.52} & \red{83.10} & \red{87.46} & \blue{90.21} & \red{86.36} & \red{85.44} & \red{64.70} & \blue{61.41} & \red{73.46} & \red{76.94} & \red{57.38} & \red{76.34} & 1,008,000  & 8\\
        & GWD-ret & 88.71 & \blue{74.91} & 46.72 & \blue{67.14} & \blue{76.05} & \blue{77.62} & 86.33 & \red{90.36} & \blue{82.10} & 83.03 & \blue{56.36} & 60.93 & 61.67 & \blue{65.82} & \blue{48.42} & \blue{71.08} & 702,000 & 4 \\
        & Ours$_{\omega=2}$ & \red{89.09} & 72.15 & \blue{46.92} & 62.22 & 75.78 & 74.70 & \blue{86.62} & 89.59 & 78.35 & \blue{83.15} & 55.83 & \red{64.01} & \blue{65.50} & 65.46 & 46.23 & 70.04 & 702,000 & 4    \end{tabular}
    \label{table:dota_gbb_results}
\end{table*}
\addtolength{\tabcolsep}{4pt}


\begin{table*}[ht!]
    \centering
    \footnotesize
    \caption{Results for OBB detection in the HRSC2016 dataset}
    \begin{tabular}{c|l|ccccc|rr}
        \\
        Model & Loss & AP50 & AP60 & AP75 & AP85 & AP50:95 & Iterations & BS\\ \hline
        \multirow{3}{*}{R-50 Retinanet} 
        & GWD-rep & \red{85.56} & \red{84.04} & \blue{60.31} & \red{17.14} & \red{52.89} & 200,000 & 8 \\
        & GWD-ret & 81.50 & 76.57 & 53.25 & 8.30 & 46.83 & 130,000 & 1 \\
        & Ours$_{\omega=0.2}$    & \blue{85.07} & \blue{83.71} & \red{61.32} & \blue{12.66} & \blue{52.55} & 130,000 & 1
        \\ \hline
        \multirow{3}{*}{R-50 R$^{3}$Det}
        & GWD-rep & \red{89.43} & \red{88.89} & \red{65.88} & \blue{15.02} & \red{56.07} & 200,000 & 8 \\
        & GWD-ret & 83.66 & 80.65 & 58.11 & 12.47 & 49.81 & 130,000 & 1 \\
        & Ours$_{\omega=0.2}$ & \blue{87.09} & \blue{85.70} & \blue{63.94} & \red{19.84} & \blue{55.61} & 130,000 & 1
    \end{tabular}
    \label{table:hrsc_gbb_results}
\end{table*}

\subsection{Limitations}
\label{sec:limitations}

Despite the promising results using GBBs and the proposed \piou-based regression loss functions, our approach presents some theoretical and practical limitations. The orientation in GBBs is inherited from the elliptical representation, so that isotropic Gaussians cannot be oriented (unlike squared-shaped OBBs). As such, the orientation of objects with a roughly square shape cannot be determined using GBBs (such as some airplanes in the DOTA dataset). On the other hand, circular objects with no obvious orientation (e.g., a ball) are encoded with an artificial orientation using OBBs, which does not happen with GBBs.

Another issue concerns very thin elongated objects, for which $a$ or $b$ might be very small (consider the axis-aligned GBB for simplicity). In these cases, the Bhatacharrya Distance might produce very large gradients for $a$ or $b$ when the compared GBBs are not aligned (see the first term in Eq.~\eqref{eq:bhatta:diagonal} and its derivative w.r.t. $a$ and $b$), which might cause instabilities in the training step and compromise convergence for this kind of objects.

\section{Conclusions}
This paper explored a fuzzy object representation using Gaussian distributions, which we called Gaussian Bounding Boxes (GBBs). Such representation allows a natural conversion to an elliptical binary mask (potentially rotated), which showed to adjust better to the object shape (segmentation mask) when compared to the popular HBBs and even OBBs. The GBB encoding also allows us to explore similarity metrics for generic probability density functions, which are strongly simplified when using Gaussian distributions. We proposed a ``Probabilistic IoU'' function (called \piou{}) based on the Hellinger Distance for matching two Gaussian distributions, which leads to differentiable loss functions involving the five parameters required to encode a 2D Gaussian (and that can also be explored for validating object detectors). We also present a scheme that allows us to seamlessly explore \piou{}-based loss functions applied to any object detector that works with either HBBs or OBBs.

Our experimental results showed that the proposed loss functions coupled to traditional HBB object detectors are competitive when compared to SOTA IoU-based regression losses, being superior when evaluated with \piou{}. When dealing with oriented object detection, the proposed loss functions showed results comparable to or better than GWD, a recent SOTA loss function proposed for GBB-like representations. However, we emphasize that the proposed  \piou{} does provide an intuitive geometrical relationship with the IoU by exploring overlap of the Gaussian PDFs, whereas an empirical non-linear mapping with additional hyper-parameters is required to adapt GWD as a regression loss. 

As future work, we intend to further investigate other statistical methods for computing the similarity or distance between PDFs, allowing other possibilities for comparing GBBs in the context of object detection.
Finally, we intend to explore a 3D version of the GBB for volumetric object detection, which seems a natural extension from the 2D version.

\section*{Supplementary Material}


%


\appendix
\section{\piou{} versus IoU}

\begin{figure}[htb]
    \centering
    \subfloat[IoU vs. \piou{} with GBBs]{
    \includegraphics[width = 7cm]{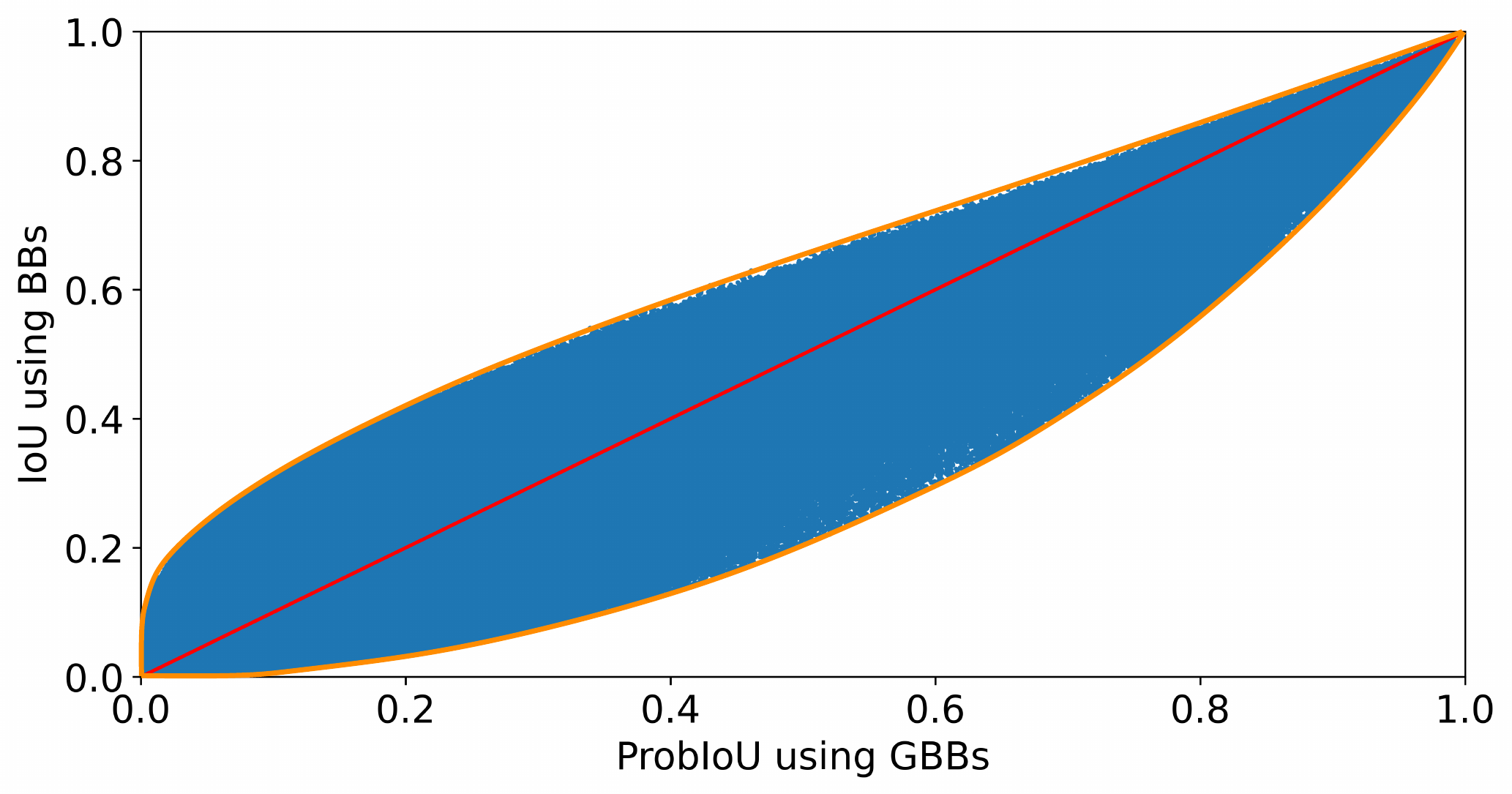}\label{fig:iou:piou}}
    \subfloat[IoU vs. \piou with HBBs]{\includegraphics[width = 7cm]{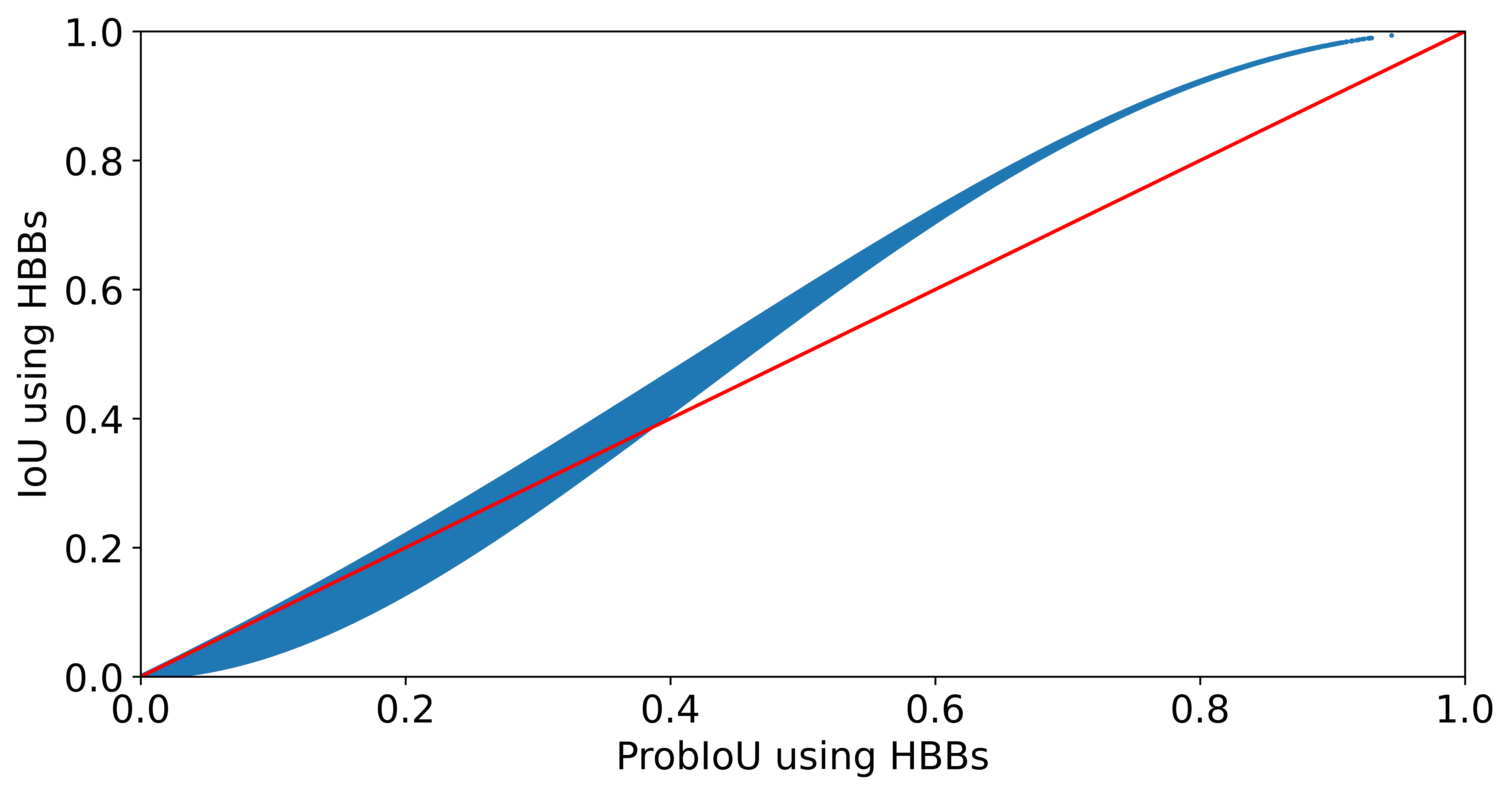}\label{fig:iou:piou:bb}}
    \caption{Empirical comparison between IoU with HBBs and \piou{} with  (a) the respective GBBs and (b) uniform PDFs. We used a set of 5,000,000 pairs of random HBBs. The red line is the identity function.}
    \label{fig:iou:piou:reps}
\end{figure}

Although the proposed \piou{} was designed based on probabilistic region overlap,
it is not possible to find a one-to-one relationship of the IoU between two generic HBBs and the proposed \piou{} between the corresponding GBBs obtained with the approach described in the paper as a fuzzy representation of the HBB. However, it is easy to see that the maximum value in both IoU and \piou{} are only obtained when the two compared representations (either HBB or GBB) are identical. To empirically illustrate the relationship between IoU and \piou{}, we generated a large set of 5,000,000 pairs of random BBs with center location and dimensions represented as random real values in the range $[0,1]$ (recall that both IoU and \piou{} are invariant to the scale changes). For each pair, we computed the IoU, the corresponding GBB representations and the \piou{} based on the GBB. Fig.~\ref{fig:iou:piou} shows a scatter plot of the obtained \piou{} and IoU pairs, and it indicates that a tighter relationship between the two similarity metrics arises as both IoU and \piou{} get closer to one. 

The proposed \piou{} based on the Hellinger Distance can also be computed using any distribution to approximate the object shape. As such, we can consider any binary representation for the shapes (HBB, OBB, or segmentation maks) as PDFs with uniform distributions (after a proper normalization). More precisely,let us consider that two object regions (e.g., HBBs) $\Omega_1$ and $\Omega_2$ are represented by PDFs $p(\bm{x})$ and $q(\bm{x})$ with uniform distributions. The Bhattacharyya Coefficient $B_C$ is given by
\begin{equation}
\label{eq:bc:prob}
    B_C(p,q) = \int_{\Omega_1\cap \Omega_2}\sqrt{p(\bm{x})q(\bm{x})}d\bm{x} = 
    \frac{\#\left(\Omega_1\cap \Omega_2\right)}{\sqrt{\#\Omega_1 \#\Omega_2}}.
\end{equation}

If we denote $N=\#\Omega_1$ and $M=\#\Omega_2$, the following inequalities are valid:
\begin{equation}
\sqrt{NM} \leq \frac{N+M}{2} \leq \max\{N,M\} \leq \#(\Omega_1 \cup \Omega_2),  
\end{equation}
which means that $B_C(p,q) \geq \text{IoU}(\Omega_1, \Omega_2)$ (not only for bounding boxes, but for any binary representation mask for the objects). Since the proposed \piou{} is computed based on the Hellinger distance, we cannot find a direct relationship between the IoU and the \piou{} computed with binary masks. However, we performed a simulated experiment by randomly generating 50,000,000 HBBs in the unit square (recall that both IoU and \piou{} are scale invariant, so that the definition of this search space is arbitrary), and computed both the IoU and the \piou{} using the HBB regions (i.e., without the conversion to GBB). The plot shown in Figure~\ref{fig:iou:piou:bb} illustrates that there is a close mapping between the IoU and the proposed \piou{} based on the Hellinger distance. However, the \piou{} with HBBs is zero for disjoint object regions, so it faces the same limitations as the IoU when using it as a regression loss.


\section{More Experimental Results}

\subsection{Comparison of segmentation masks with HBBs, OBBs and GBB-induced ellipses}
In the paper we show that computing the covariance matrix from the segmentation masks in the COCO 2017 dataset yields elliptical regions that are closer to the mask (in terms of IoU) than traditional representations such as HBBs and even OBBs. Here we expand the analysis and show a per-category comparison.

\begin{figure}
    \centering
    \includegraphics[width = 6.5cm]{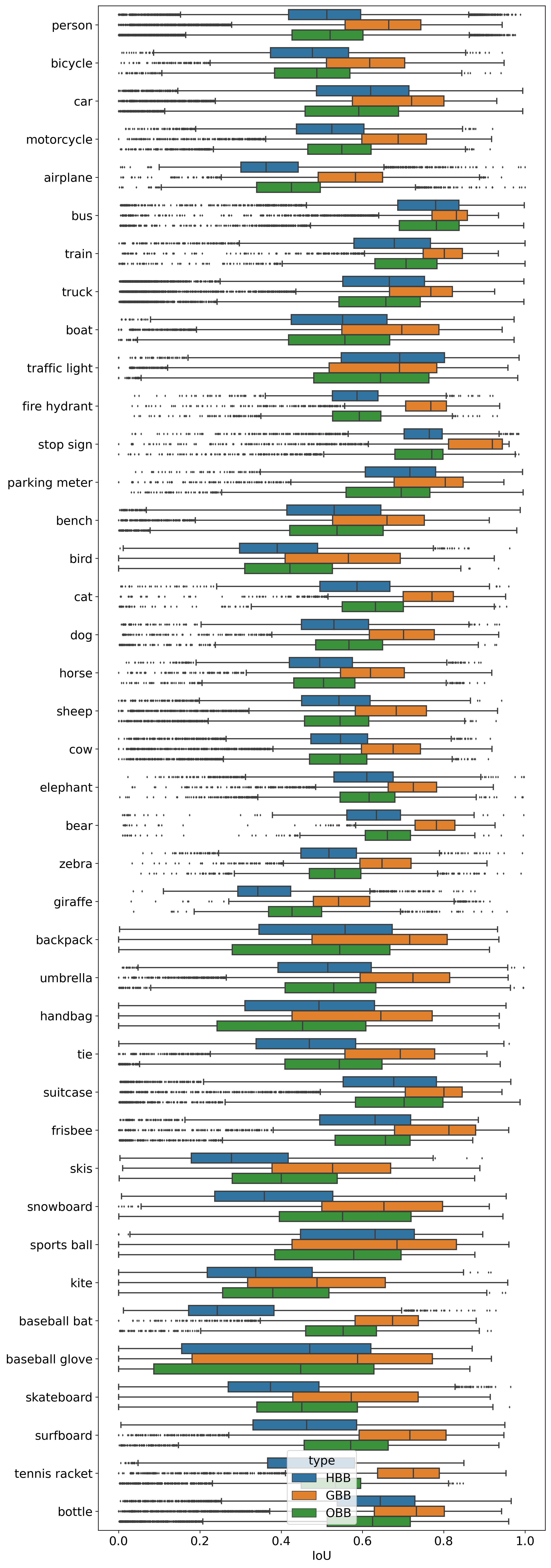}
    \includegraphics[width = 6.5cm]{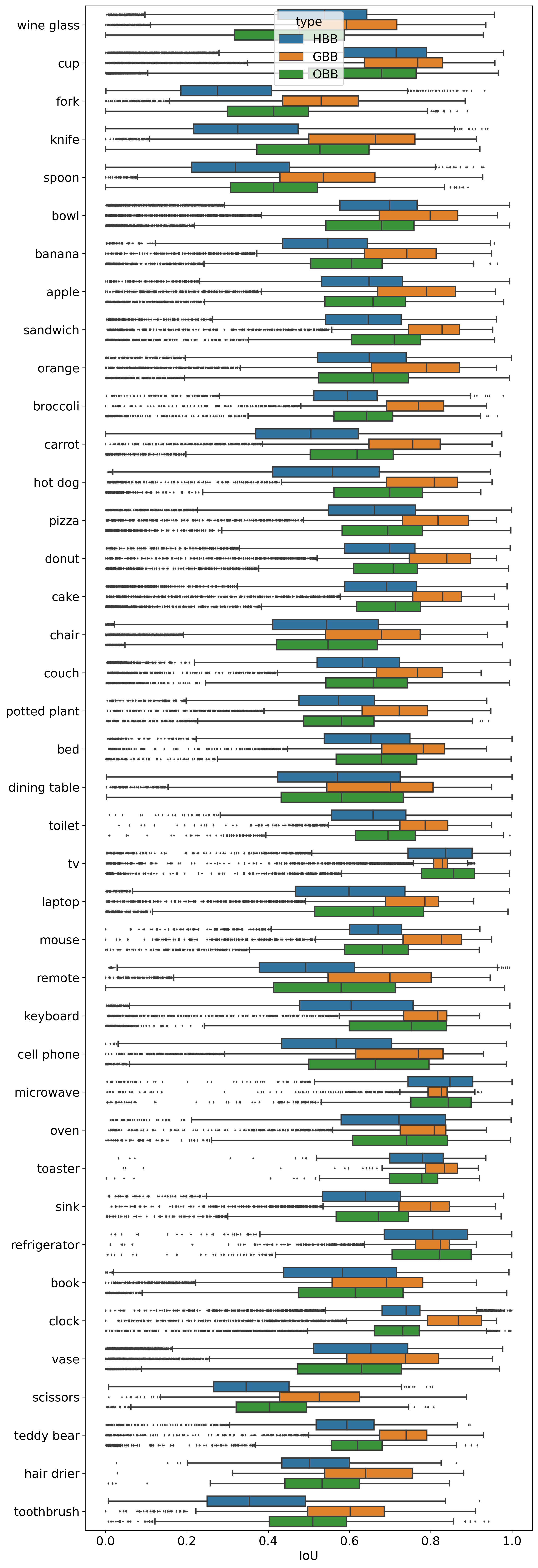}
    \caption{IoU between segmentation masks and HBBs, OBBs and GBBs for each category of the COCO 2017 train set.}
    \label{fig:iou_comparison}
\end{figure}

The distribution of the IoU values between the segmentation mask and the three representations (HBBs, OBBs and GBBs) for all categories in the COCO 2017 dataset are shown as boxplots in Fig.~\ref{fig:iou_comparison},  and it shows that using GBBs yields higher median IoU values than HBBs for the vast majority (77 out of 80) of the categories. The exceptions were \verb'traffic light' and \verb'microwave', for with the HBB representation was the best, and \verb'tv', for which OBB was the winner.

Our analysis also shows that the median IoU values with HBBs are smaller than 0.5 in 31 out of the 80 categories in the dataset. This means that we use an ideal object detector that produces the ground-truth bounding box and evaluate it with the segmentation mask using the traditional IoU threshold of 0.5, it would produce wrong detections for more than half of the samples in these 31 categories. In some of them, such as \verb'skis', \verb'baseball bat' and \verb'fork', the median HBB IoU was smaller than 0.25. When using GBBs, the number of categories for which the median IoU is smaller than 0.5 drops to only one: \verb'kite'. The categories for which GBB presented the largest median IoU values ($> 0.85$) were \verb'stop sign' and \verb'clock', which is actually expected due to the circular nature of these objects.
Finally, the IoU values obtained with OBBs fell below 0.5 in 13 categories, and the only category that produced a median IoU value over 0.85 was \verb'tv'.

\begin{figure}[htb]
    \centering
    \includegraphics[height = 2cm]{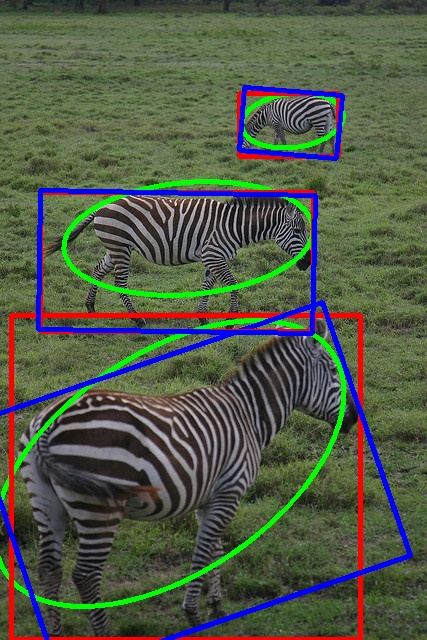}~\includegraphics[height = 2cm]{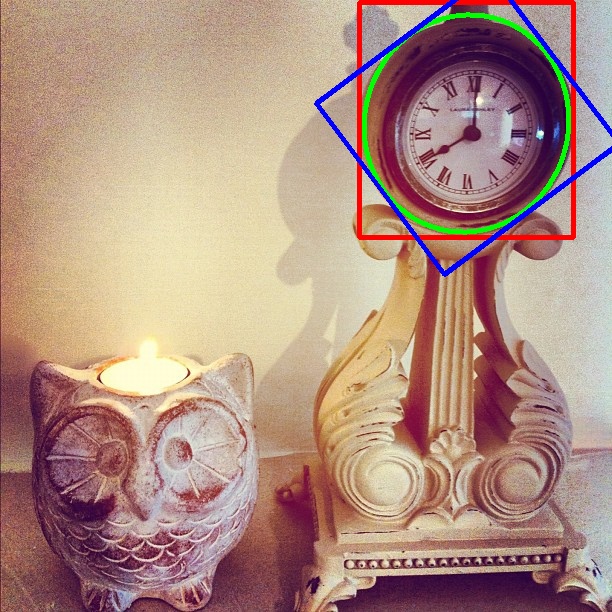}~\includegraphics[height = 2cm]{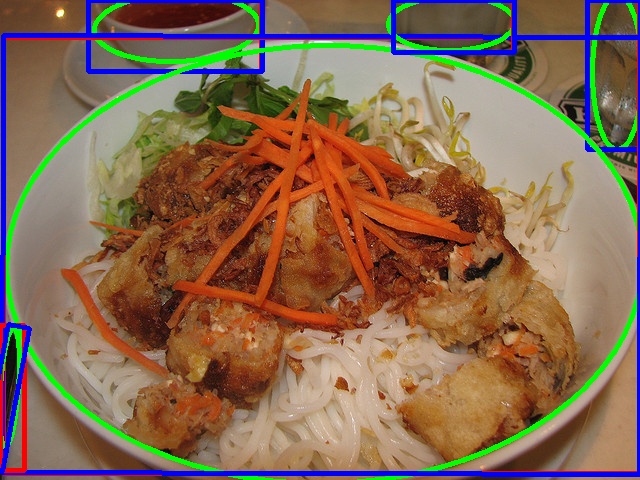}~\includegraphics[height = 2cm]{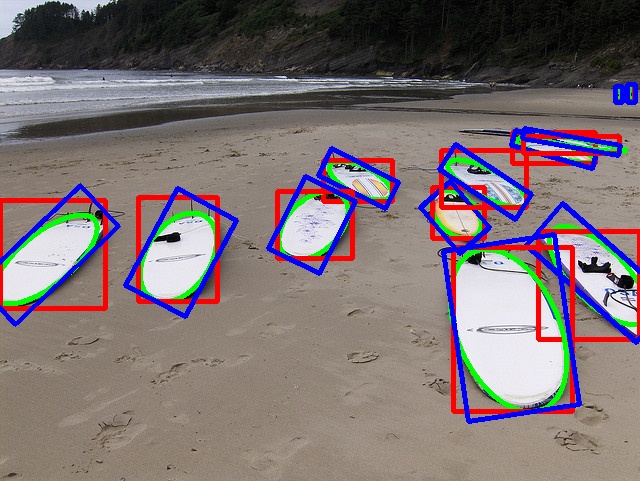}~\includegraphics[height = 2cm]{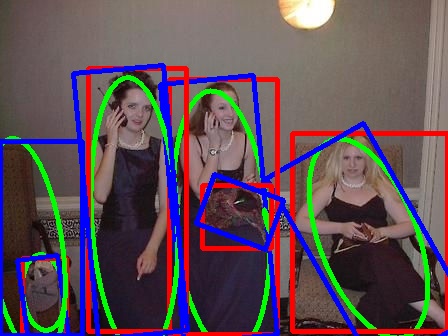}~\includegraphics[height = 2cm]{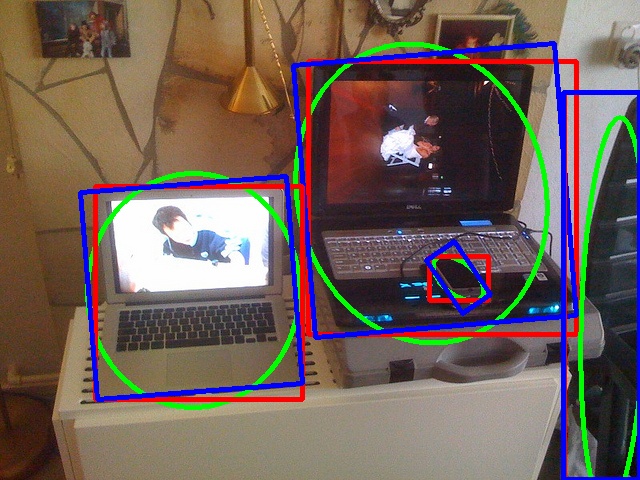}
    \\
        \includegraphics[height = 2cm]{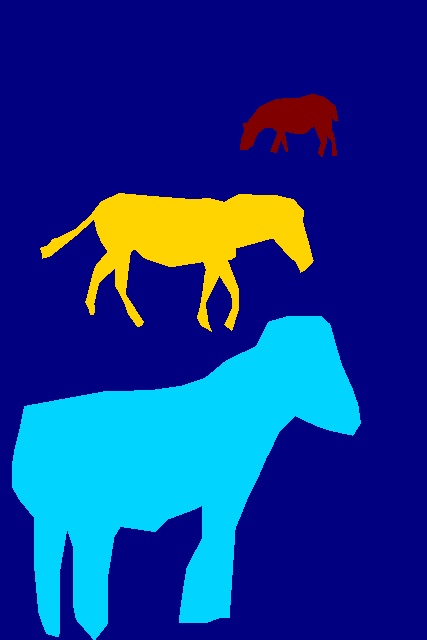}~\includegraphics[height = 2cm]{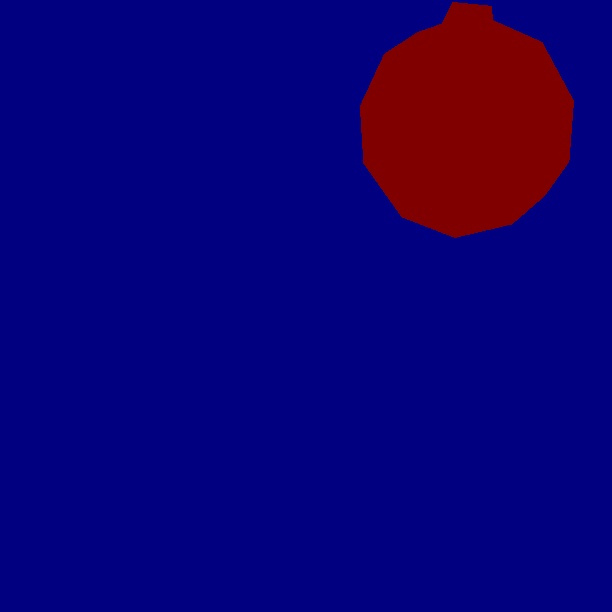}~\includegraphics[height = 2cm]{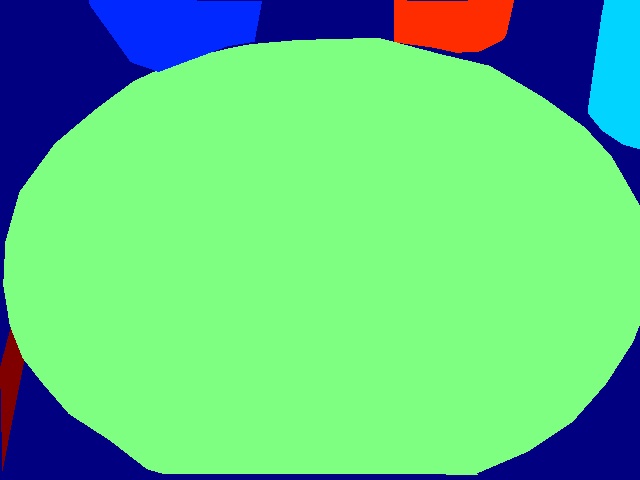}~\includegraphics[height = 2cm]{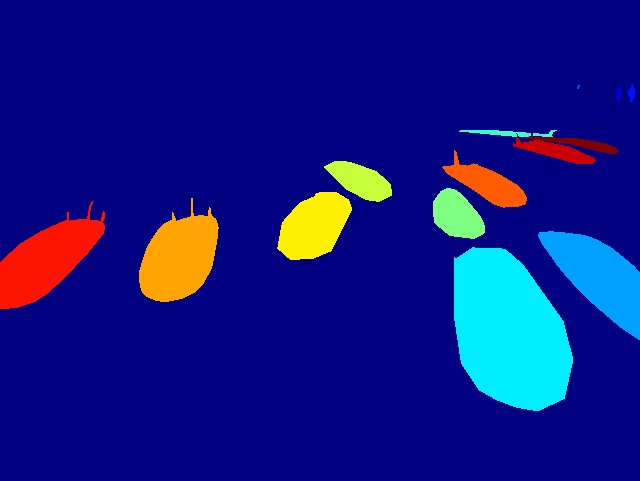}~\includegraphics[height = 2cm]{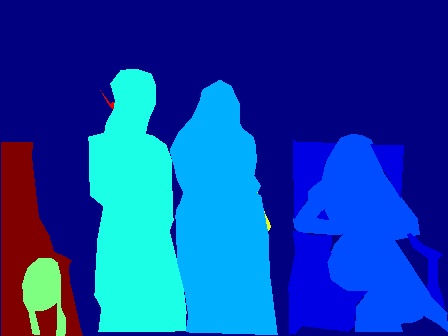}~\includegraphics[height = 2cm]{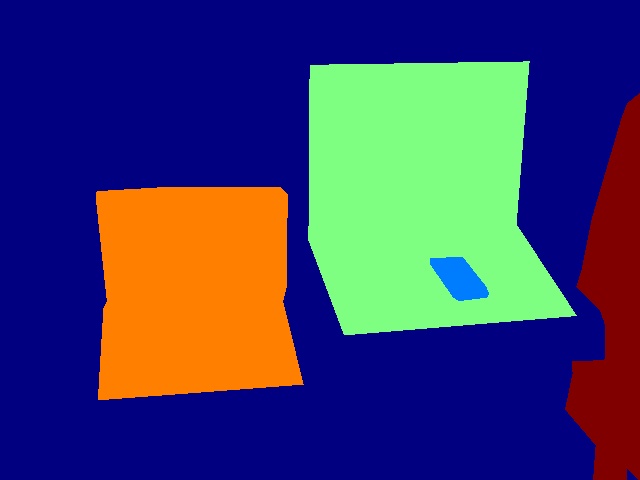}
    \caption{Top: object representations as HBBs (red), OBBs (blue) and GBB-induced ellipses (green). Bottom: corresponding segmentation masks.}
    \label{fig:iou:examples}
\end{figure}

Considering the IoU of all instances (over 700K, regardless of the category), the median IoU values for HBBs, OBBs and GBBs are, respectively, 0.55, 0.56, and 0.70. This means that using GBBs yields an IoU improvement of around 28\% over both HBB and OBB representations.  In this individual analysis, 42.9 \% of the samples present a BB IoU smaller than 0.5, compared to 20.7\% using GBBs. These results indicate that just adding rotation to BBs still might not produce substantial overlap with the GT segmentation masks for generic objects (such as the COCO dataset), since the shapes might not be rectangular. On the other hand, our elliptical representations seem to be a better match. 
Figure~\ref{fig:iou:examples} shows more visual examples comparing the different object representations (top) with the annotated segmentation masks (bottom). In most images, the elliptical regions better match the segmentation mask (corroborating the analysis above), and exceptions are objects with a naturally rectangular shape, such as computers (see the images on the right).

\subsection{Experimental Setup and Visual results for HBB and OBB Object Detection}

Here we provide additional details about our experiments, as well as the used hardware in each experiment and visual results HBB and OBB object detection. The code used for training all the models reported in the paper can be found in \url{https://github.com/ProbIOU}. 

\subsubsection{HBB object detection}
The hardware setup used for training the HBB models on the PASCAL VOC 2007 dataset~\cite{VOC2007} were:
\begin{itemize}
    \item \textbf{Efficient Det D0~\cite{tan2020efficientdet}:} we used 2 Nvidia RTX 2080 GPUs with 12 GB of VRAM. Training each experiment takes approximately 13 hours.
    \item \textbf{SSD300~\cite{Liu:ECCV:2016}:} We used 1 Titan XP with 12GB of VRAM. Training each experiment takes approximately 18 hours. 
\end{itemize}

\begin{figure}[!h]
    \centering
        \subfloat[IoU = 87.80\\ ProbIoU = 93.26 ]{\includegraphics[width=62px,height = 92px]{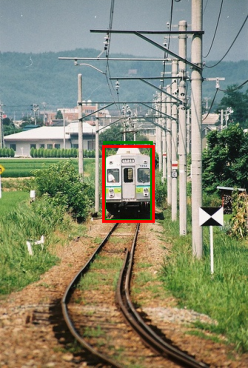}}~
        \subfloat[IoU =  93.05 \\ ProbIoU = 94.78   ]{\includegraphics[width=62px,height = 92px]{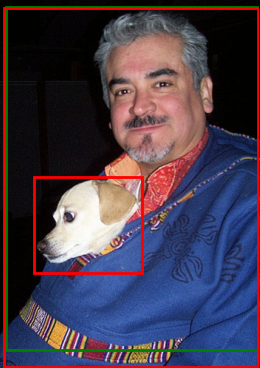}}~
        \subfloat[IoU = 83.18 \\ ProbIoU = 90.61 ]{\includegraphics[width=62px,height = 92px]{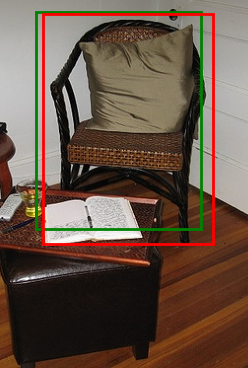}}~
        \subfloat[IoU = 80.09 \\ ProbIoU = 87.40  ]{\includegraphics[width=62px,height = 92px]{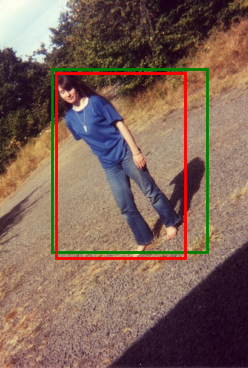}} ~\subfloat[IoU = 57.79 \\ ProbIoU = 70.02  ]{\includegraphics[width=62px,height = 92px]{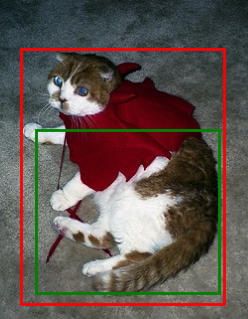}}
        ~\subfloat[IoU = 84.84 \\ ProbIoU = 90.86 ]{\includegraphics[width=62px,height = 92px]{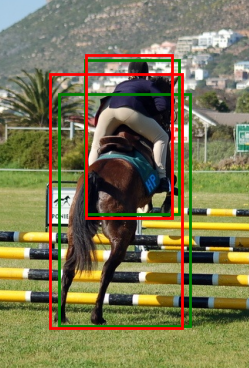}}
        \\
        \subfloat[IoU = 18.11 \\ ProbIoU = 27.10 ]{\includegraphics[width=97px,height = 62px]{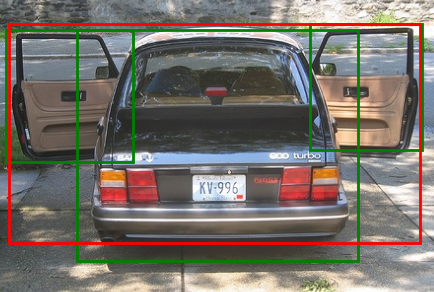}}~\subfloat[IoU = 8.24 \\ ProbIoU = 19.27 ]{\includegraphics[width=97px,height = 62px]{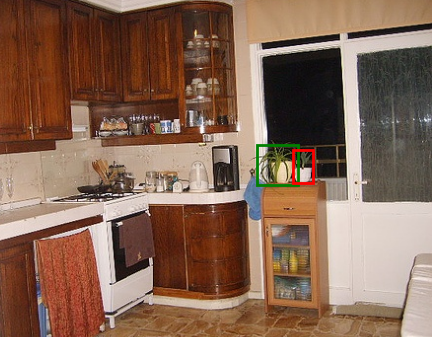}}~\subfloat[IoU = 75.00 \\ ProbIoU = 96.43 ]{\includegraphics[width=97px,height = 62px]{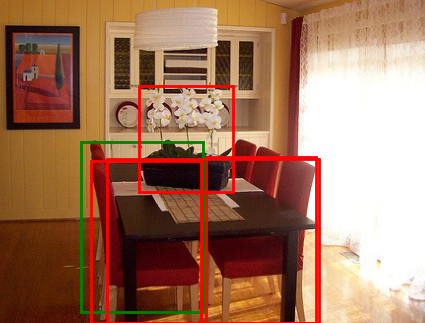}}~\subfloat[IoU = 77.36 \\ ProbIoU = 86.60 ]{\includegraphics[width=97px,height = 62px]{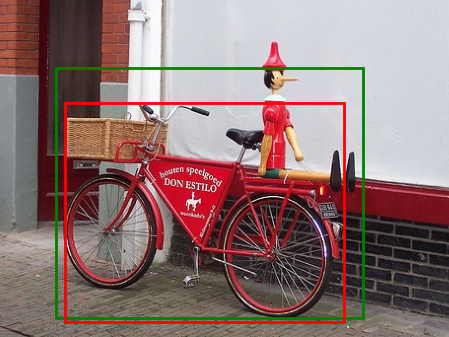}}
    \caption{OBB detection results (in green) for VOC dataset using SSD trained with our \piou{}-based loss functions (ground truth HBBs shown in red).} 
    \label{fig:projected_images}
\end{figure}

For the sake of illustration, we show in Figure~\ref{fig:projected_images} some detection results (in red, while the GT HBBs are shown in green) using SSD~\cite{Liu:ECCV:2016} trained with our \piou{}-based loss ($10L_2\rightarrow L_1$), along with the IoU and \piou{} validation scores (when more than one object is detected, these metrics related to the detection with the highest score). We can observe that wrong detections (low IoU or undetected objects) usually involve overlap or similar objects present in the image. We can also note that the \piou{} evaluation value correlates with the IoU, being usually a bit higher (as indicated in Figure~\ref{fig:iou:piou}). 

\subsubsection{OBB object detection}
The hardware setup for training the OBB models for the tested datasets were:
\begin{itemize}
    \item \textbf{R-50 Retinanet~\cite{lin2017focal}} \begin{itemize}
        \item DOTA v1~\cite{Xia:2018:CVPR}: we used 2 Nvidia RTX 2080 GPUs with 12 GB of VRAM. Training each experiment takes approximately 28 hours.
        \item HRSC2016~\cite{liu:icpram:2017}: we used 1 Nvidia RTX 2080 GPU with 12 GB of VRAM. Training each experiment takes approximately 8 hours.
    \end{itemize}
    
    \item \textbf{R-50 R$^3$Det~\cite{yang2021r3det}} \begin{itemize}
        \item DOTA v1~\cite{Xia:2018:CVPR}: We used 4 Tesla T4 GPUs with 16GB of VRAM. Training each experiment takes approximately 2 days. 
        \item HRSC2016~\cite{liu:icpram:2017}: we used 1 Nvidia RTX 2080 GPUs with 12 GB of VRAM. Training each experiment takes approximately 12 hours.
    \end{itemize}
\end{itemize}

\begin{figure}
    \centering
\includegraphics[height = 3cm]{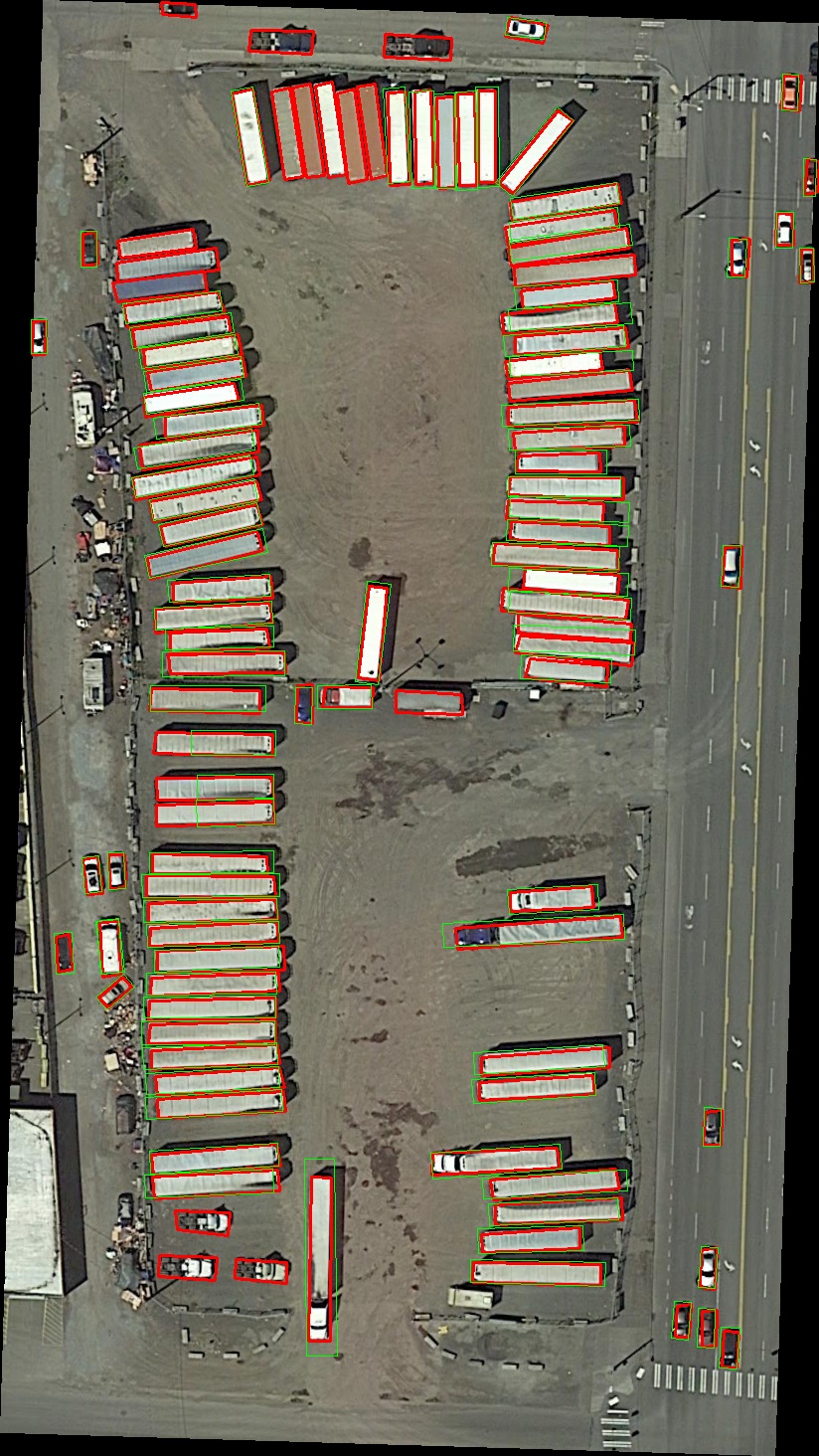}
\includegraphics[height = 3cm]{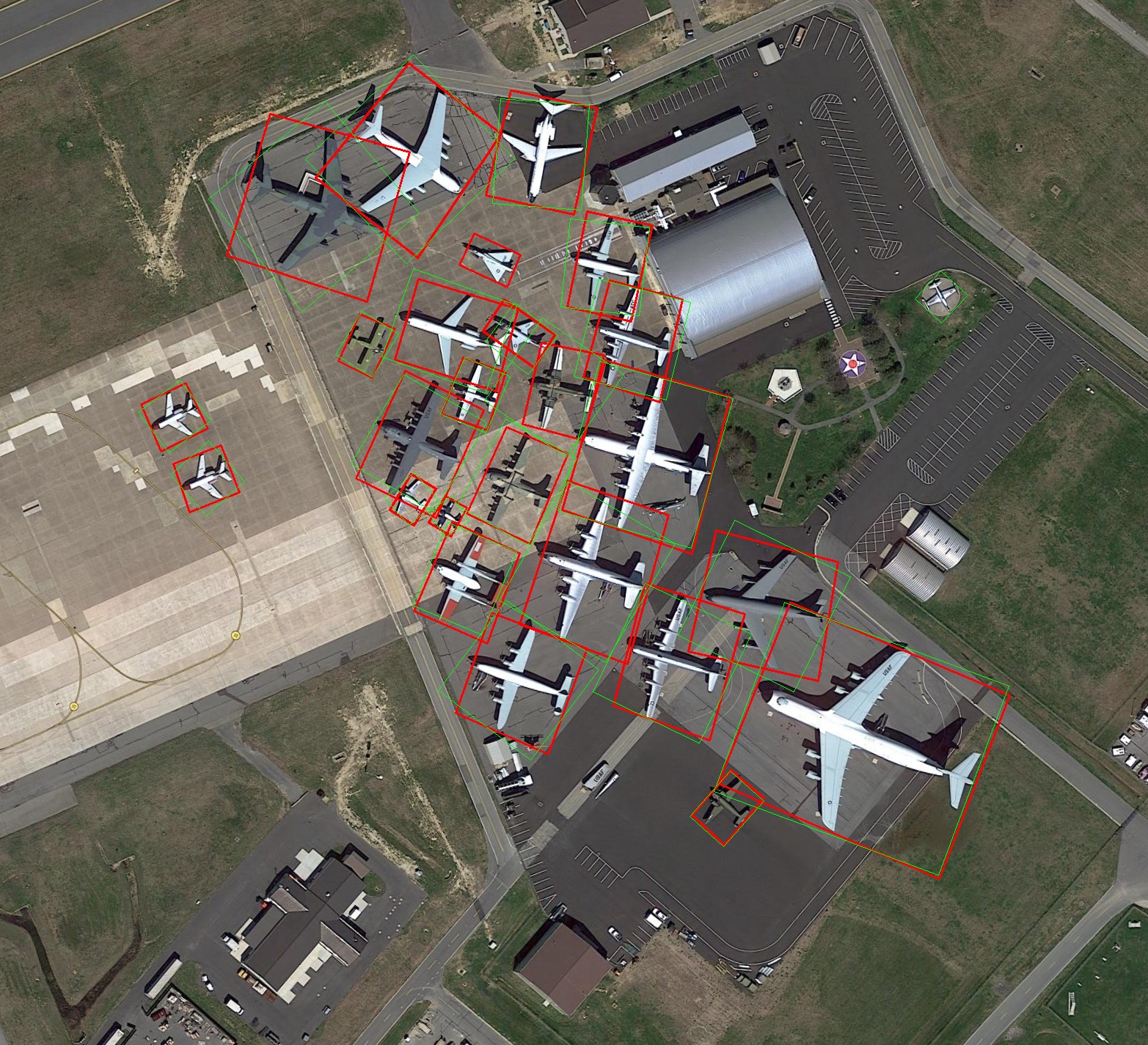}
\includegraphics[height = 3cm]{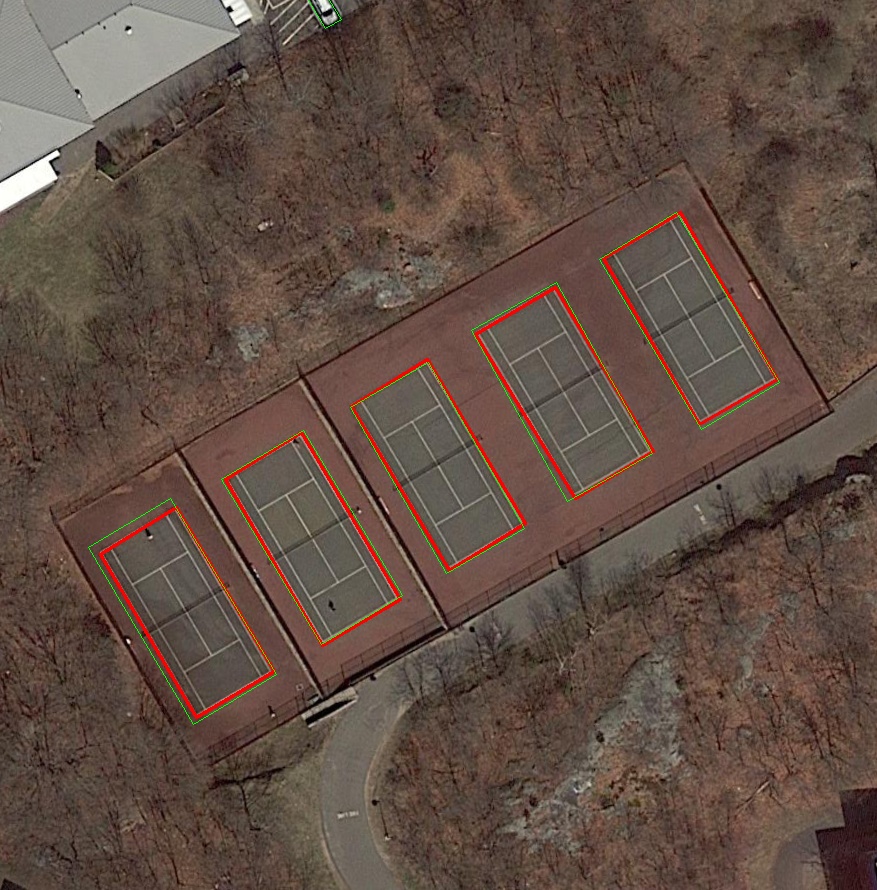}
\includegraphics[height = 3cm]{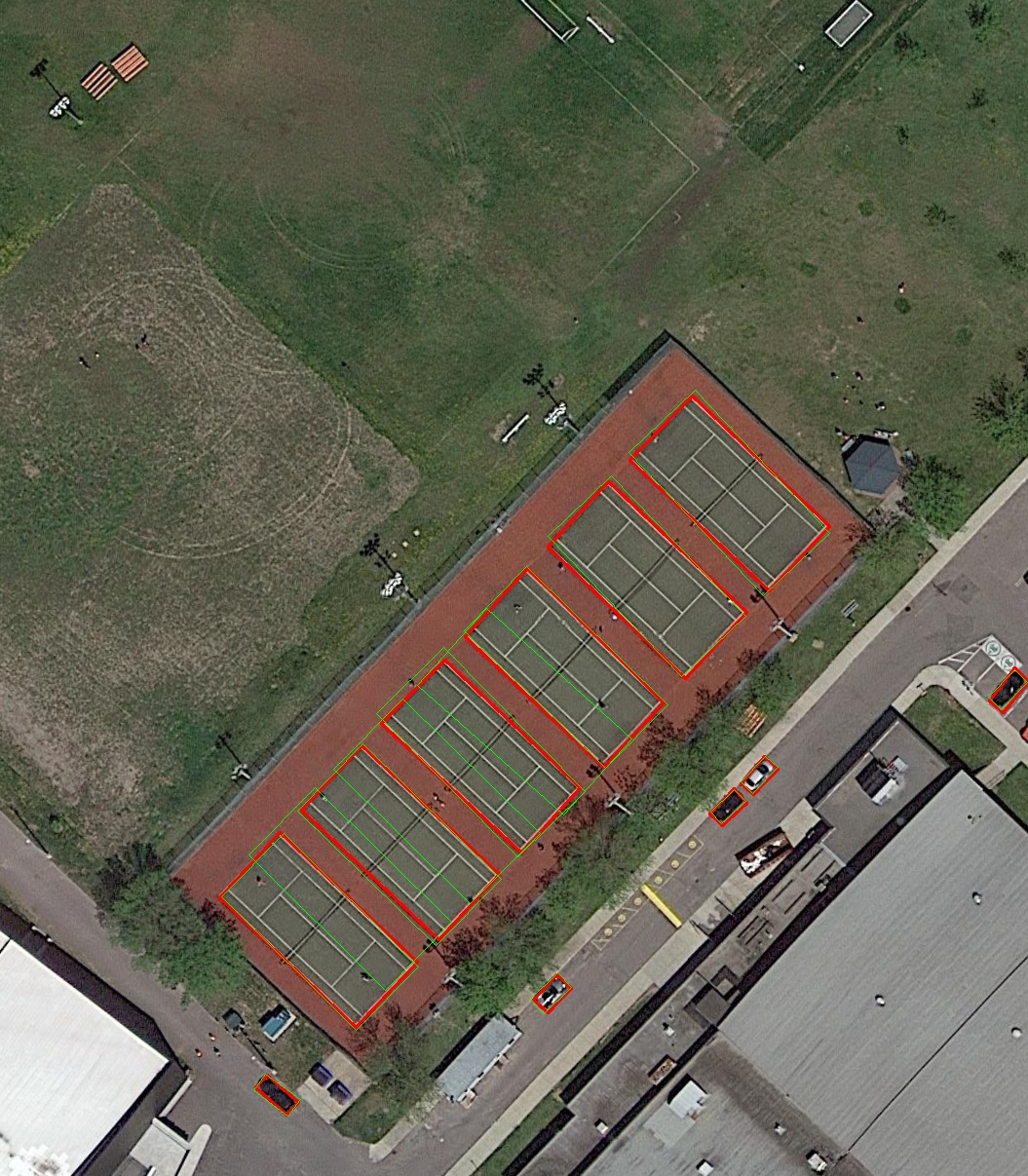}
\includegraphics[height = 3cm]{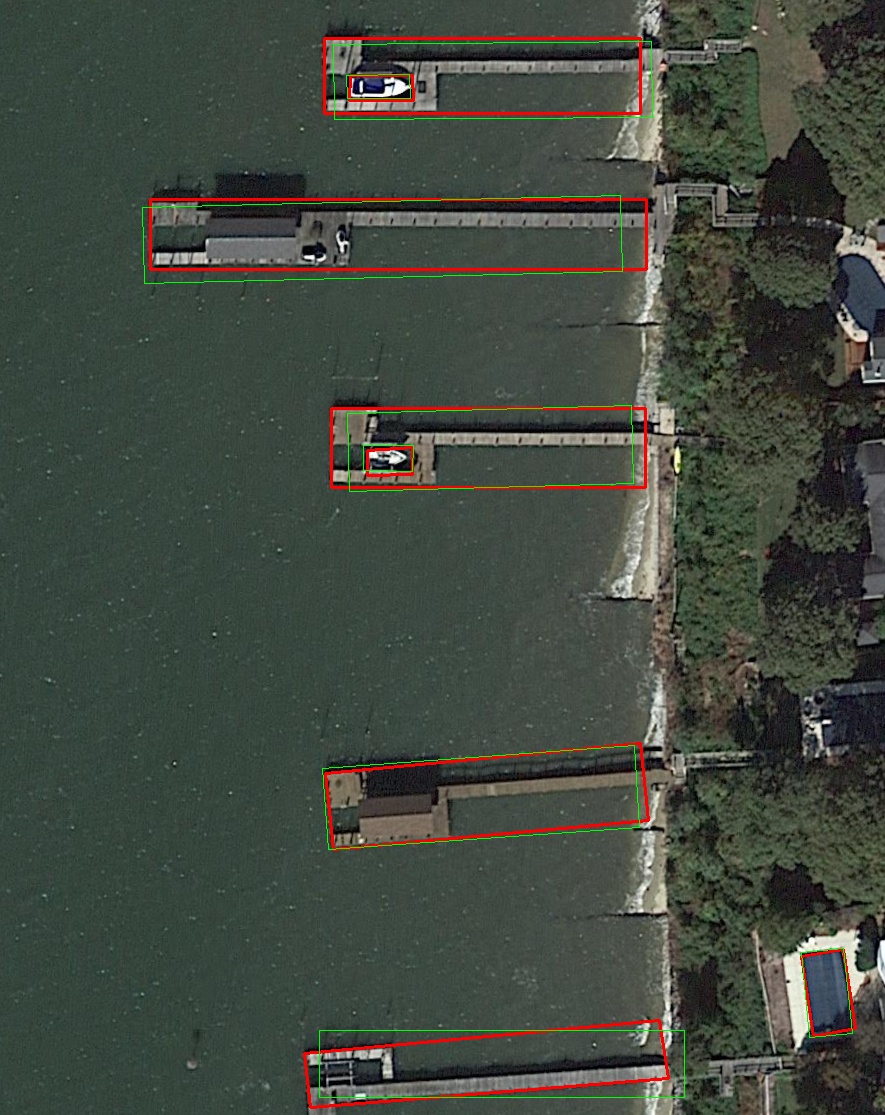} \\
\includegraphics[height = 3cm]{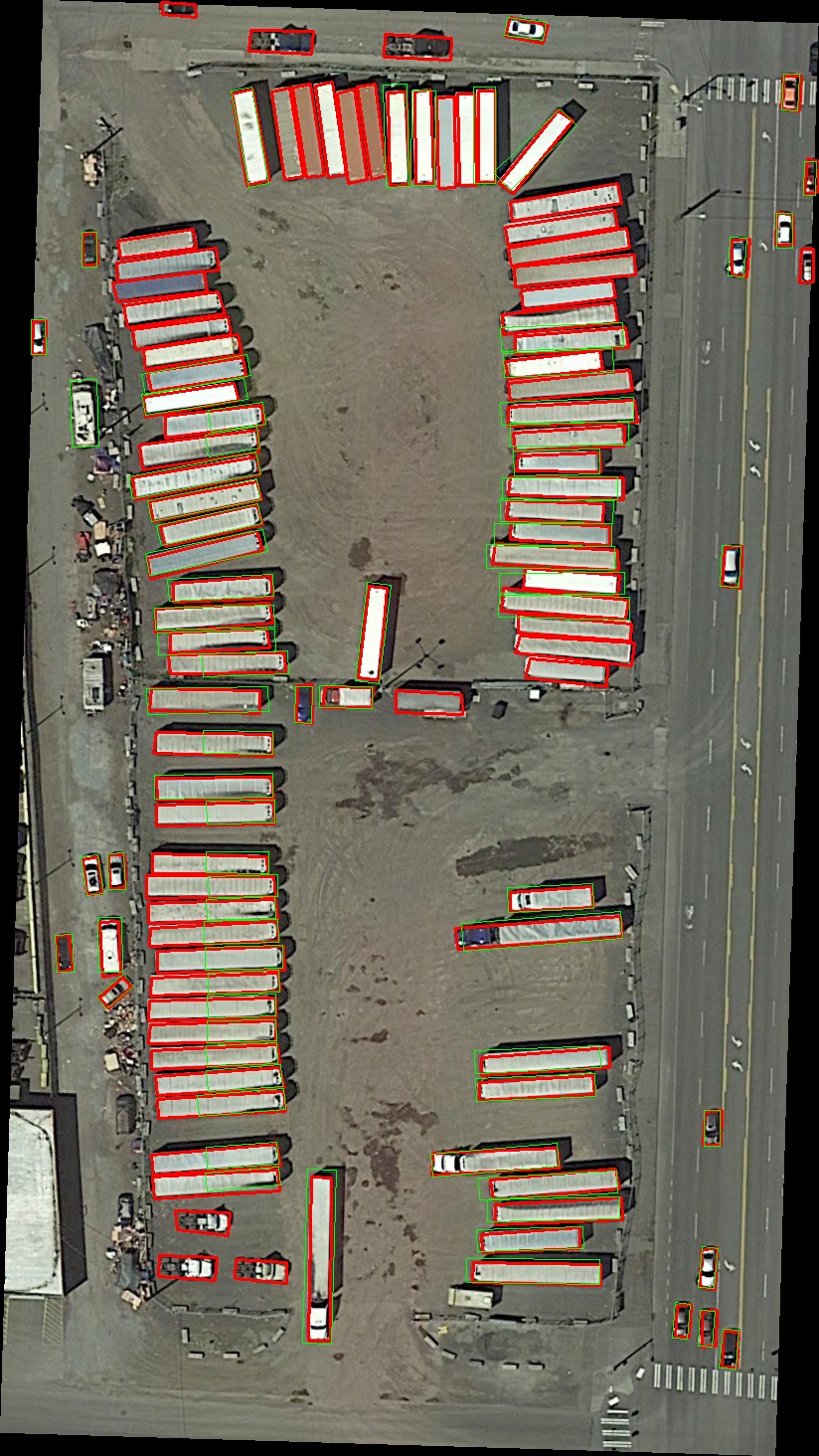}
\includegraphics[height = 3cm]{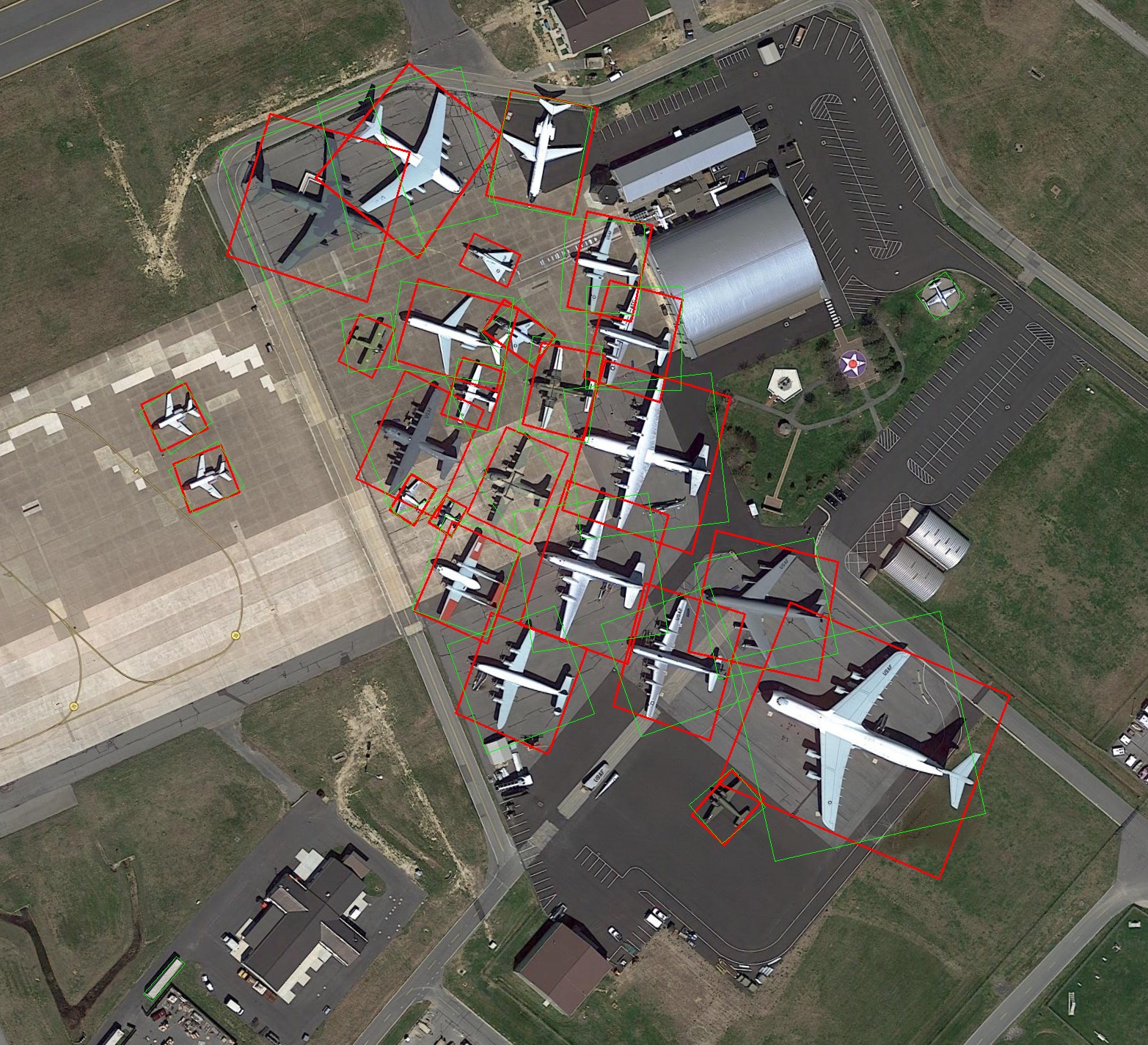}
\includegraphics[height = 3cm]{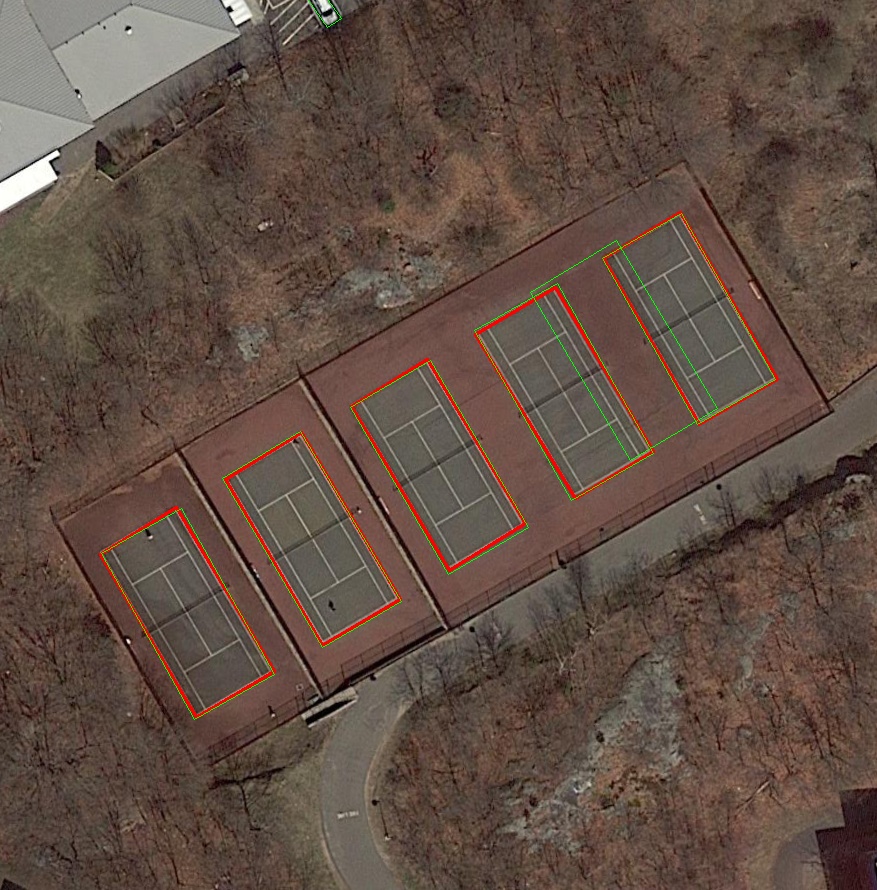}
\includegraphics[height = 3cm]{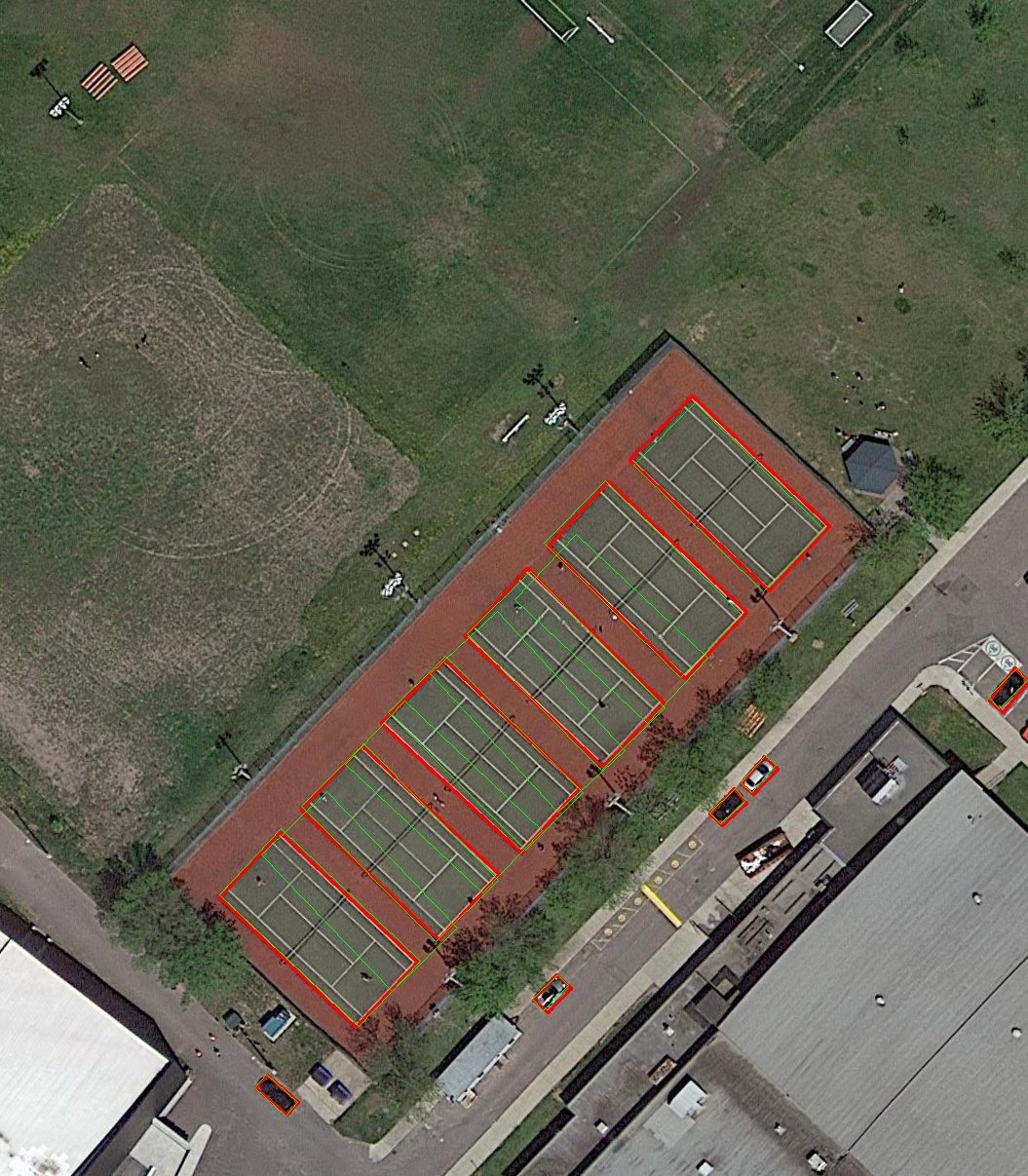}
\includegraphics[height = 3cm]{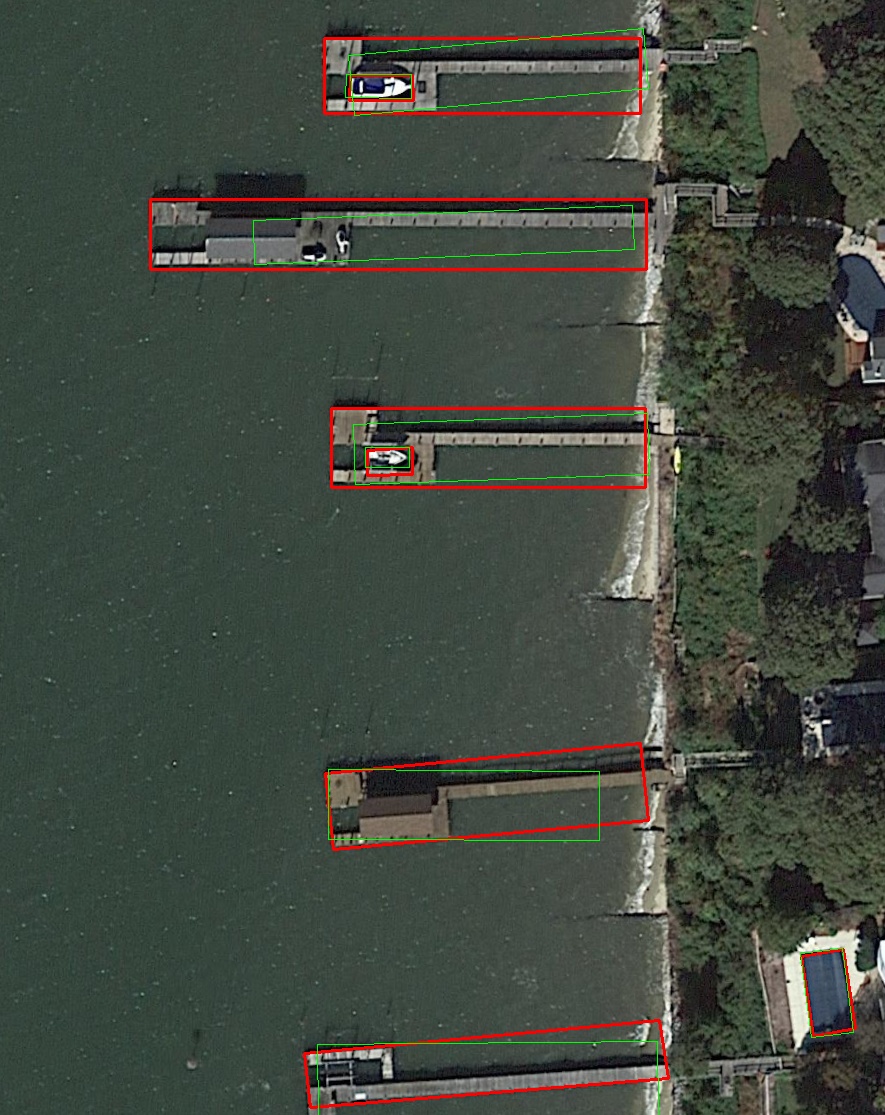} 
    \caption{Detection results (in green) for the DOTA v1 dataset using R-50 RetinaNet (ground truth OBBs shown in red). Top: results using our loss. Bottom: results using GWD.}
    \label{fig:images:dota}
\end{figure}

A few visual examples of OBBs using the proposed loss ($10\mathcal{L}_2\rightarrow2\mathcal{L}_1$) for the DOTA v1 dataset are shown in Figure~\ref{fig:images:dota} on the top row (for the sake of comparison, results using GWD shown in the bottom). We can observe that most objects are correctly detected, but the orientation for roughly square-shaped annotations (such as some planes in the second image) contains larger errors (for both our loss and GWD). This behavior is actually expected since the underlying Gaussian distribution in these cases is roughly isotropic, and varying the orientation does not change the distribution significantly (in the limit case of an isotropic distribution, the orientation simply does not exist).

\section{Ablation study}

The only hyper-parameters in our method are the weights $\omega_1$ and $\omega_2$ responsible for balancing the localization loss using $\mathcal{L}_1$ or $\mathcal{L}_2$, respectively, with the classification loss. In all experiments, the final loss was computed through
\begin{equation}
    \mathcal{L} = \mathcal{L}_{clas} + \omega_j\mathcal{L}_{j}, ~~j\in\{1,2\},
\end{equation}
recalling that we suggest a two-stage training process starting with $\mathcal{L}_2$ and then switching to $\mathcal{L}_1$. Next we show some ablations studies by using only $\mathcal{L}_1$ or $\mathcal{L}_2$, and also by changing the values of $\omega_1, \omega_2$, noting that in the paper notation the definition of a single value $\omega$ means that $\omega_1 = \omega$ and $\omega_2 = 5\omega$. In all experiments for the same classifier and dataset, the total number of iterations was the same as reported in the paper.


\subsection{Results for HBB detection}

The ablation results for HBB detection using EfficientDet D0~\cite{tan2020efficientdet} and SSD300~\cite{Liu:ECCV:2016} are shown in Tables~\ref{table:efficientdet:ablation} and~\ref{table:ssd:ablation}, respectively. In all reported results, the  symbol ``$\rightarrow$'' means that the training started with the loss function indicated on the left (with the corresponding weight), and then switched to the loss on the right (with the respective weight) in the middle of training. We can see that (at least within the parameter combinations that we tested), the best results for EfficientDet D0 using both IoU and \piou{} as evaluation metrics were achieved with $\omega_1=10$, $\omega_2=2$, and that switching from the $\mathcal{L}_2$ to $\mathcal{L}_1$ in the middle of training in general improved the final results. For SSD~\cite{Liu:ECCV:2016}, we can note that the combination of parameters that yields the highest AP using \piou{} ($\omega_1 = 1$, $\omega_2 = 10$) are not the same that produce the best AP scores using IoU ($\omega_1 = 2$, $\omega_2 = 10$). Although we have not performed an exhaustive combination of the parameters, we can see that an adequate choice to balance the classification and localization loss terms is important for achieving better results (as noted in~\cite{zheng2020distance} for the CIoU and DIoU loss functions).

\begin{table*}[ht!]
    \centering
    \caption{Results for HBB with EfficientDet D0 in the PASCAL VOC 2007 dataset}
    \begin{tabular}{c|cc|cc}
        \\
        \multirow{2}{*}{Loss} & \multicolumn{2}{c}{AP} &  \multicolumn{2}{|c}{AP75}  \\  \cline{2-5}
        & \textbf{IoU}  & \textbf{ProbIoU} & \textbf{IoU} & \textbf{ProbIoU} \\ \hline
        $1 \boldsymbol{\mathcal{L}_2}$ & 33.88 & 49.59 & 30.33 & 56.79 \\
        $1 \boldsymbol{\mathcal{L}_1}$ & 37.14 & 51.01 & 37.11 & 58.11 \\
        $1 \boldsymbol{\mathcal{L}_2}\rightarrow1 \boldsymbol{\mathcal{L}_1}$ & 39.99 & 53.99 & 41.09 & 60.75 \\
        $1 \boldsymbol{\mathcal{L}_2}\rightarrow2 \boldsymbol{\mathcal{L}_1}$ & 40.43 & 54.42 & 41.41 & 61.60 \\
        $1 \boldsymbol{\mathcal{L}_2}\rightarrow4 \boldsymbol{\mathcal{L}_1}$ & \blue{42.40} & \blue{56.37} & \blue{43.93} & 63.58 \\
        $20 \boldsymbol{\mathcal{L}_2}\rightarrow2 \boldsymbol{\mathcal{L}_1}$ & 41.59 & 56.06 & 42.46 & \blue{63.79} \\
        $10 \boldsymbol{\mathcal{L}_2}\rightarrow2 \boldsymbol{\mathcal{L}_1}$ & \red{42.60} & \red{56.76} & \red{44.24} & \red{64.15}
    \end{tabular}
    \label{table:efficientdet:ablation}
\end{table*}

\begin{table*}[ht!]
    \centering
    \caption{Results for HBB with SSD300 in the PASCAL VOC 2007 dataset}
    \begin{tabular}{c|cc|cc}
        \\
        \multirow{2}{*}{Loss} & \multicolumn{2}{c}{AP} &  \multicolumn{2}{|c}{AP75}  \\  \cline{2-5}
        & \textbf{IoU}  & \textbf{ProbIoU} & \textbf{IoU} & \textbf{ProbIoU} \\ \hline
        $1 \boldsymbol{\mathcal{L}_2}$ & 41.10 & 62.93 & 41.81 & 71.15 \\
        $1 \boldsymbol{\mathcal{L}_1}$ & 41.22 & 63.74 & 41.66 & \red{72.54} \\
        $2 \boldsymbol{\mathcal{L}_1}$ & 41.42 & 63.17 & 43.22 & 71.60 \\
        $1 \boldsymbol{\mathcal{L}_2}\rightarrow1 \boldsymbol{\mathcal{L}_1}$ & 41.68 & 63.37 & 42.32 & 72.29 \\
        $10 \boldsymbol{\mathcal{L}_2}\rightarrow1 \boldsymbol{\mathcal{L}_1}$ & \blue{41.89} & \red{64.16} & \blue{43.00} & \blue{72.39} \\
        $10 \boldsymbol{\mathcal{L}_2}\rightarrow2 \boldsymbol{\mathcal{L}_1}$ & \red{42.11} & \blue{63.39} & \red{43.89} & 71.63 \\
    \end{tabular}
        \label{table:ssd:ablation}
\end{table*}

\subsection{Results for OBB detection}

Due to time/hardware constraints noted above for DOTA v1, we limited the ablation study for OBB object detection using the HRSC2016 
dataset~\cite{liu:icpram:2017} with the R-50 RetinaNet detector. It is interesting to note that by changing the hyper-parameters, we can achieve even better results than those reported in the paper ($\omega = 0.2$). For example, using $5 \boldsymbol{\mathcal{L}_2}\rightarrow1 \boldsymbol{\mathcal{L}_1}$ yields an AP50:95 value of 54.23, which is even higher than the accuracy (52.89) reported in~\cite{yang2021rethinking} using larger batch sizes and more iterations. We also noted that using $\mathcal{L}_1$ alone presented very good results as well (the best ones in 4 out of 5 metrics). As we will discuss next, the rationale for starting with $\mathcal{L}_2$ is to avoid vanishing gradients potentially produced when using $\mathcal{L}_1$ in the beginning of the training step, and then fine-tune GBB matching with $\mathcal{L}_1$. We believe that in this ablation study the potential problem of vanishing gradients did not happen when using $\mathcal{L}_1$ from the start, yielding very good AP values only with $\mathcal{L}_1$.



\begin{table*}[ht!]
    \centering
    \footnotesize
    \caption{Results for OBB detection in the HRSC2016 dataset with R-50 RetinaNet}
    \begin{tabular}{c|ccccc}
        \\
        Loss & AP50 & AP60 & AP75 & AP85 & AP50:95\\ \hline
        $1 \boldsymbol{\mathcal{L}_2}$ & 85.88 & 83.08 & 57.96 & 10.43 & 50.05 \\
        $1 \boldsymbol{\mathcal{L}_1}$ & \red{87.03} & \red{85.31} & \red{64.07} & \blue{15.86} & \red{54.48} \\
        $10 \boldsymbol{\mathcal{L}_2}\rightarrow2 \boldsymbol{\mathcal{L}_1}$ & 84.45 & 80.73 & 49.03 & 9.53 & 46.76 \\
        $1 \boldsymbol{\mathcal{L}_2}\rightarrow0.2 \boldsymbol{\mathcal{L}_1}$ & 85.07 & 83.71 & 61.32 & 12.66 & 52.55 \\
        $5 \boldsymbol{\mathcal{L}_2}\rightarrow1 \boldsymbol{\mathcal{L}_1}$ & \blue{86.81} & \blue{85.06} & \blue{62.65} & \red{18.71} & \blue{54.23}
    \end{tabular}
    \label{table:hrsc_gbb_results}
\end{table*}


\subsection{Convergence and Potential Limitations}

\begin{figure}[ht!]
    \centering
    \small
    \subfloat[$\textrm{IoU}=0$; $\textrm{ProbIoU}=7.48$\\ $\|\nabla{\mathcal{L}_2}\|=10.6663$; \\ $\|\nabla{\mathcal{L}_1}\|=0.4583$]{\framebox{\includegraphics[width=4.5cm]{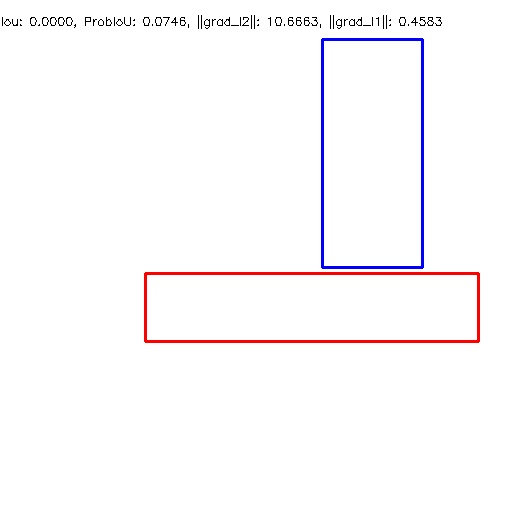}}}
    ~\subfloat[$\textrm{IoU}=0$; $\textrm{ProbIoU}=0.02$\\ $\|\nabla{\mathcal{L}_2}\|=105.3814$; \\ $\|\nabla{\mathcal{L}_1}\|=0.0000$]{\framebox{\includegraphics[width=4.5cm]{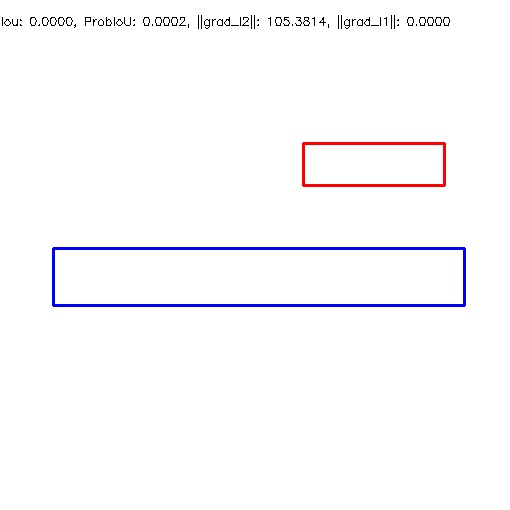}}}~
    \subfloat[$\textrm{IoU}=74.30$; $\textrm{ProbIoU}=86.98$\\ $\|\nabla{\mathcal{L}_2}\|=0.2362$; \\ $\|\nabla{\mathcal{L}_1}\|=0.8629$]{\framebox{\includegraphics[width=4.5cm]{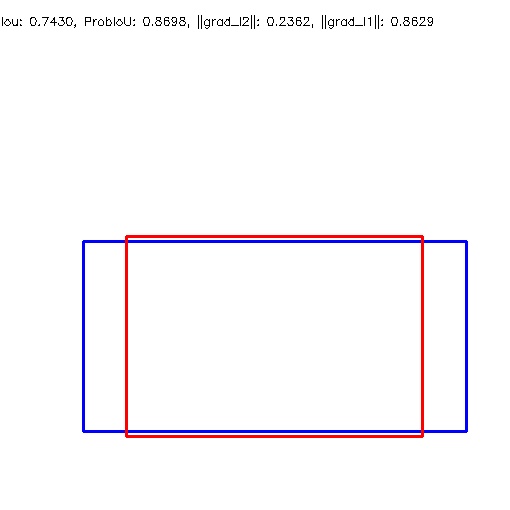}}}
    \caption{Illustration of the IoU, \piou{} and the gradient magnitudes of $\mathcal{L}_1$ and $\mathcal{L}_2$ for the GBBs associated with the shown HBBs.}
    \label{fig:gradient}
\end{figure}

Using $\mathcal{L}_1 =1 - \piou$ as the localization loss follows a recent trend of using IoU-based functions as done in~\cite{rezatofighi2019generalized, zheng2020distance}. However, $\mathcal{L}_1$ might rapidly tend to one when the compared GBBs are far apart from each other (which is typically the case when you start training the model, as shown in Figure~\ref{fig:gradient}b), and hence the gradient might be very close to zero. In these cases, convergence using  $\mathcal{L}_1$ directly might be very slow or generate local minima.  Fortunately,  $\mathcal{L}_2$ does not suffer from this limitation, since it is not upper-bounded. On the other hand, the gradient of $\mathcal{L}_2$ might get small as the GBBs get closer to each other (as shown in Figure~\ref{fig:gradient}c), whereas $\mathcal{L}_1$ yields larger values that tend to produce a tighter fit. As an example, Figure~\ref{fig:gradient} shows pairs of HBBs in different configurations, along with the values for the IoU, \piou{}, $\| \nabla\mathcal{L}_2\|$ and $\| \nabla\mathcal{L}_1\|$. Analytical expressions for the gradients of $\mathcal{L}_1$ and $\mathcal{L}_2$ are provided in the Appendix.

For the reasons mentioned above, starting the models training with the $\mathcal{L}_2$ loss and then switching for the $\mathcal{L}_1$ usually produces better results than training only with one of them. However, the existing problem of correctly setting the weights for the losses (as described in~\cite{zheng2020distance}) arises twice, i.e., the models are more sensitive to the choice of the weight. Moreover, although switching the losses in the middle of training with a pre-defined number of iterations  (mainly for its simplicity) yields good results, a more suitable method would be to wait for the $\mathcal{L}_2$ loss to stabilize before switching it to  $\mathcal{L}_1$ (e.g., monitoring the $\mathcal{L}_2$ validation loss and switch when it stops to improve), and then automatically finding a suitable weight $\omega_1$ based on the stabilization value of $\mathcal{L}_2$ to start the second training stage with a coherent localization loss value  $\omega_1\mathcal{L}_1$.

Finally, we note that  $\| \nabla\mathcal{L}_2\|$ and $\| \nabla\mathcal{L}_1\|$ might produce very large gradients for small or elongated boxes. For the case of HBBs, these gradients are computed explicitly in the Appendix, and we can see that if the width $W_2$ or height $H_2$ of the annotated box is small (and hence so are the corresponding parameters $W_1$ and $H_1$ of the regressed box, assuming convergence near the end of training), then $\| \nabla\mathcal{L}_2\|$ might be large according to Eq.~\eqref{eq:gradient:l2}. Also, Eq.~\eqref{eq:gradient:l1} indicates that $\| \nabla\mathcal{L}_1\|$ tends to be even higher, which might cause instabilities in the training process.

\section{Regressing Gaussian Bounding Box (GBB) parameters directly}

As discussed in the paper, the objects are represented in a fuzzy way through 2D Gaussian distributions. They are uniquely described by the mean vector $\bm{\mu} = (x_0, y_0)^T$ and covariance matrix $\Sigma$, which is a symmetric positive-definite matrix with three degrees of freedom. It can be expressed directly as the matrix elements $a,b,c$, or as the decorrelated variances $a',b'$ and the rotation angle $\theta$, which are related through
\begin{equation}
\label{cov:rotation}
    \Sigma = \begin{bmatrix}
    a & c\\
    c & b 
    \end{bmatrix} =  R_{\theta}\begin{bmatrix}
    a' & 0\\
    0 & b' 
    \end{bmatrix}R_{\theta}^T = \begin{bmatrix}a' \cos^{2}\theta + b' \sin^{2}\theta & \frac{1}{2}\left(a' - b'\right) \sin2 \theta \\
    \frac{1}{2}\left(a' - b'\right) \sin2 \theta  & a' \sin^{2}\theta  + b' \cos^{2}\theta\end{bmatrix},
\end{equation}

In the paper, we showed results with the latter representation, since it allows an easier adaptation to existing object detectors that work with Oriented Bounding Boxes (OBBs). Now, we present an alternative for regressing the parameters $(a,b,c)$ directly, recalling that positive-defineteness implies that $a>0$ and $\det \Sigma = ab - c^2 >0$ (which also implies that $b > 0$).  These constraints can be imposed to the newtork output in several ways, and in this paper we  propose the following strategy:
\begin{enumerate}
    \item Regress $c$ without constraints, meaning that it can be an output parameter of the network with linear activation
    \item Instead of regressing $a,b > 0$ directly, regress a new set of parameters $\alpha,\beta$ without constraints (linear activation).
    \item Define $a = e^\alpha >0$, and $b = c^2/a + e^\beta = e^{-\alpha}c^2 + e^\beta$, so that $ab >  c^2$ is satisfied $\forall c,\alpha,\beta$. To avoid numerical instabilities (underflow or overflow) caused by the exponential mapping, clamping might be applied to $\alpha,\beta$
\end{enumerate}

As a proof-of-concept (POC) for regressing the covariance matrix coefficients with the approach described above, we tackled the problem of glioblastoma cell detection using visible light microscopy images. The shape and size of glioblastoma cells can vary considerably, but they have a mostly elongated shape. As such, the proposed GBB-induced ellipses are a suitable representation, and we trained a classifier (with only one class) using an object detector similar to ~\cite{montazzolli:tits:2021}, which was design to detect license plates represented as quadrilaterals. The detector
employs residual layers as a backbone and detection head presents, for each spatial position of the last layer,
a probability $p$ of encountering an object (cell), and five (intermediate) parameters related to the GBB:  $x_0,y_0,\alpha,\beta,c$. We used a linear activation functions for all localization parameters, and applied the non-linear mappings $a=e^{\alpha}$ and $b = e^{-\alpha}c^2 + e^\beta$ to obtain the final GBB parameters. For this POC, we selected $\mathcal{L}_2$ as the localization loss.

\begin{figure}[htb]
    \centering
    \subfloat[Input image]{
    \includegraphics[width = 6.95cm]{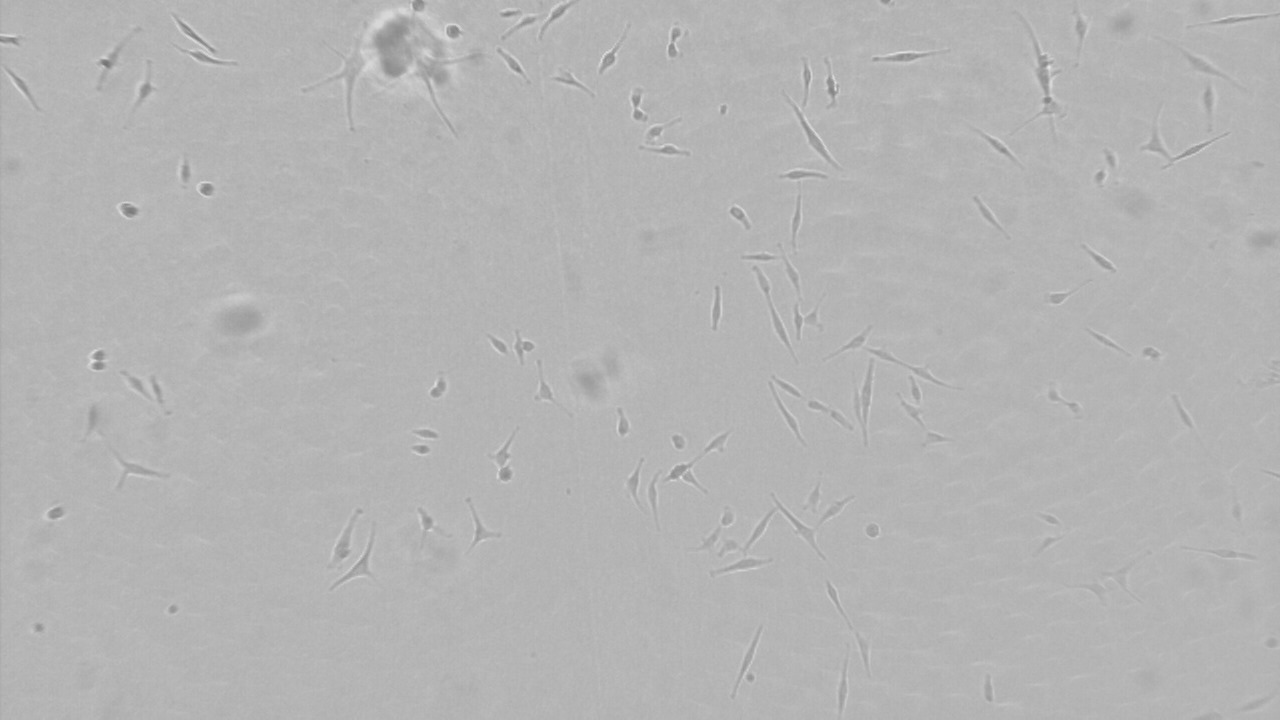}\label{cell:input}}~
    \subfloat[Detection results]{\includegraphics[width = 6.95cm]{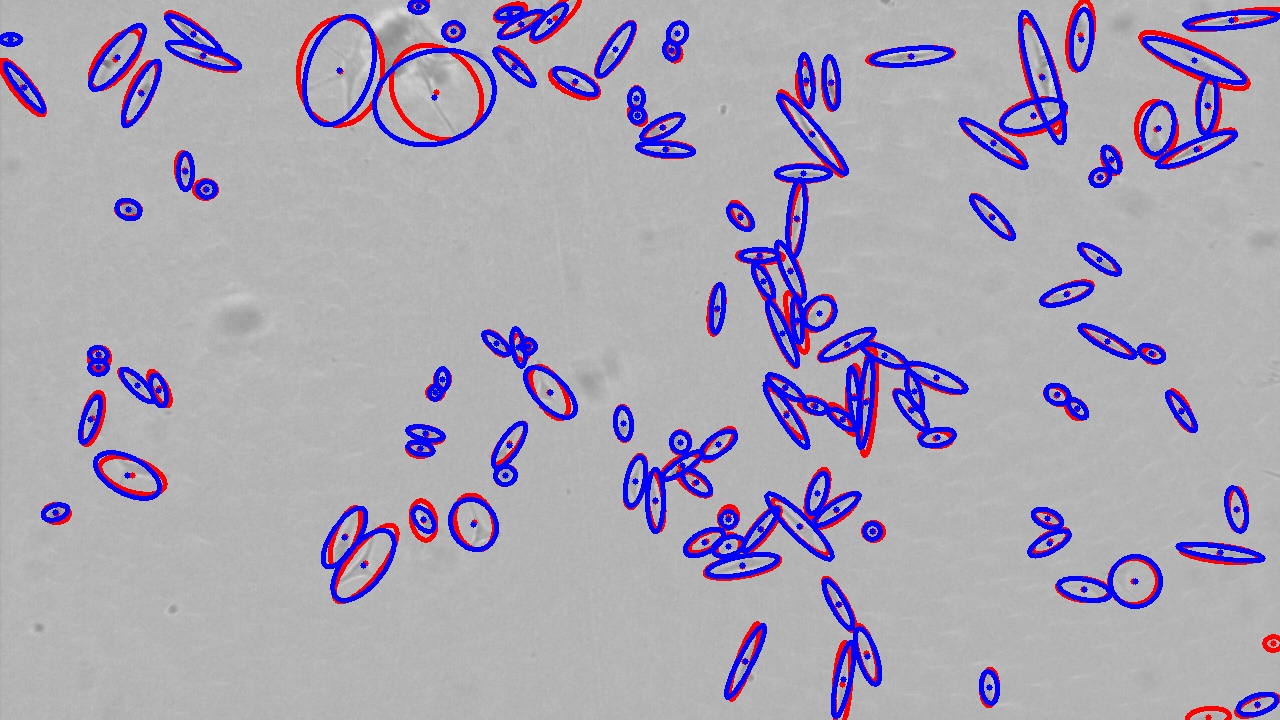}\label{cell:results}}
    \caption{Example of glioblastoma cell detection using GBB regression. (a) Original image. (b) Processed image, with detections in red and GT annotations in blue.}
    \label{fig:glioblastoma}
\end{figure}

We used a private dataset (under construction) that contains at the moment only 55  images ($1280 \times 720$) with a total of 2,937 annotated cells.  Figure~\ref{cell:input} shows an example of glioblastoma cell image, while the annotated cells (blue) and the detection results (red) are shown in Figure~\ref{cell:results}. Although these results are preliminary, they indicate the potential of direct GBB regression for object detection. The main advantage of this approach is that it does not involve any angular parameter at all, and hence there are no parametrization issues (as discussed in~\cite{yang2021rethinking} for OBBs).

\section*{Appendix}

\textbf{Relation between the GBB-induced ellipses and OBBs:}
In the paper we proposed to use the squared Mahalanobis distance
 $d^2(\bm{x}) = \left(\bm{x} - \bm{\mu} \right)^T\Sigma^{-1}\left(\bm{x} - \bm{\mu} \right)$ to obtain a binary elliptical mask from a fuzzy GBB representation. 
 Given an area threshold $0<\tau < 1$, we can set $r$ such that $F_{\chi^2}(r^2) = \tau$, where $F_{\chi^2}$ is the Cumulative Distribution Function (CDF) of the chi-squared distribution. This strategy ensures that a fraction $\tau$ of the 2D Gaussian distribution lies within the chosen ellipse. Our default choice $r = \sqrt{\frac{12}{\pi}}$ leads to an ellipse with the same area as an OBB with the same covariance matrix, as we will show next.
 
As discussed in the paper, an OBB with dimensions $H\times W$ and orientation $\theta$ induce a covariance matrix $\Sigma$ with eigenvalues $\lambda_1 = W^2/12$ and $\lambda_2 = H^2/12$. Hence, the eigenvalues of $\Sigma^{-1}$ are $\lambda_1'=1/\lambda_1=12/W^2$
and $\lambda_2'=1/\lambda_2=12/H^2$. The semi-axes of the ellipse $d^2(\bm{x}) = r^2$ are then given by $a'=r/\sqrt{\lambda_1'} = \sqrt{\pi}/W$ and $b'= r/\sqrt{\lambda_2'}=\sqrt{\pi}/H$, so that the area of the ellipse is given by $a'b'\pi =HW $, which is exactly the area of the corresponding OBB.

\textbf{Gradients of $\mathcal{L}_1$ and $\mathcal{L}_2$ for axis-aligned GBBs:} As shown in the paper, an axis-aligned GBB computed from a regressed HBB is parametrized as $\bm{p}_{G} =  (x_1,y_1,a_1,b_1) =  \bm{f}(\bm{p}_{B}) = (x_1,y_1,W_1^2/12,H_1^2/12)$, where $\bm{p}_{B} = (x,y,W,H)$ encodes the center $(x_1,y_1)$, width $W_1$ and height $H_1$ of an HBB, and a similar operation can be applied to the GT annotation:  $\bm{q}_{G} =  (x_2,y_2,a_2,b_2) = \bm{f}(\bm{q}_{B})$, so that a GBB-based loss function $\mathcal{L}_G$ can be used.

The Bhatthacharyya Distance $B_D(\bm{p}_G, \bm{q}_G)$, which is our choice for $\mathcal{L}_2$, reduces to (discarding constant terms)
\begin{equation}
B_D(\bm{p}_G, \bm{q}_G) = \frac{1}{4}\left(\frac{(x_1-x_2)^2}{a_1+a_2} +\frac{(y_1-y_2)^2}{b_1+b_2} \right) 
+\frac{1}{2}\ln\left( (a_1+a_2)(b_1+ b_2) \right) 
-
\frac{1}{4}\ln\left(a_1a_2b_1b_2\right). 
\label{eq:bhatta:diagonal}
\end{equation}

Hence, the gradient $\partial \mathcal{L}_{2}/\partial \bm{p}_{B}  = \partial B_D/\partial/ \bm{p}_{B}$, required to regress the HBB parameters directly using the GBB loss,  depends on both the Jacobian of $\bm{f}$ (which is diagonal and very easy to obtain) and the partial derivatives of the Bhattacharrya Distance $\partial B_D/\partial \bm{p}_G$. More precisely, it is given by 

\begin{align}
\label{eq:gradient:l2}
    \frac{\partial B_D}{\partial \bm{p}_B} &= \begin{bmatrix}
    1 & 0 & 0 & 0 \\
    0 & 1 & 0 & 0 \\
    0 & 0 & W_1/6 & 0 \\
    0 & 0 & 0 &  H_1/6 
\end{bmatrix}
\left(
    \begin{array}{c}
         \frac{1}{2} \frac{x_1-x_2}{a_1+a_2}\\
         \frac{1}{2} \frac{y_1-y_2}{b_1+b_2}\\
        \frac{1}{4}\left(\frac{a_1-a_2}{a_1(a_1+a_2)} -\frac{(x_1 - x_2)^2}{(a_1 + a_2)^2}\right)
 \\
\frac{1}{4}\left(\frac{b_1-b_2}{b_2(b_1+b_2)}-\frac{(y_1 - y_2)^2}{(b_1 + b_2)^2}\right)
    \end{array}
    \right) \\
    &=
    \left(
    \begin{array}{c}
    6\frac{x_1-x_2}{W_1^2+W_2^2}\\
    6\frac{y_1-y_2}{H_1^2+H_2^2} \\
    \left( \frac{W_1^2-W_2^2}{2W_1(W_1^2+W_2^2)} -6\frac{W_1(x_1-x_2)^2}{(W_1^2+W_2^2)^2}    \right)\\
    6\frac{y_1-y_2}{H_1^2+H_2^2} \\
    \left( \frac{H_1^2-H_2^2}{2H_1(H_1^2+H_2^2)} -6\frac{H_1(y_1-y_2)^2}{(H_1^2+H_2^2)^2}    \right)
\end{array}
\right),
\end{align}
yielding a simple closed-form expression that could be used directly instead of computing the gradients numerically. We can also observe that 
$\partial \mathcal{L}_{2}/\partial \bm{p}_{B} =  \bm{0} \Leftrightarrow \partial \mathcal{L}_{2}/\partial \bm{p}_{G} =  \bm{0} \Leftrightarrow \bm{p}_G = \bm{q}_G \Leftrightarrow \bm{p}_B = \bm{q}_B$, which means that we cannot have vanishing gradients during training when using $\mathcal{L}_2$ as the localization loss. 

Since we have $\mathcal{L}_1 = \sqrt{1 - \text{exp}(-B_D)} = \sqrt{1 - \text{exp}(-\mathcal{L}_2)}$, we can compute  $\partial \mathcal{L}_{1}/\partial \bm{p}_{B}$ using the chain rule to obtain
\begin{equation}
\label{eq:gradient:l1}
\frac{\partial \mathcal{L}_{1}}{\partial \bm{p}_{B}} = \frac{\exp(-\mathcal{L}_2)}{2\sqrt{1-\exp(-\mathcal{L}_2})} \frac{\partial\mathcal{L}_{2}}{\partial \bm{p}_{B} },
\end{equation}
which also satisfies $\partial \mathcal{L}_{1}/\partial \bm{p}_{B} =  \bm{0} \Leftrightarrow \partial \mathcal{L}_{1}/\partial \bm{p}_{G} =  \bm{0} \Leftrightarrow \bm{p}_G = \bm{q}_G \Leftrightarrow \bm{p}_B = \bm{q}_B$. Hence, both $\mathcal{L}_{1}$ and $\mathcal{L}_{2}$ are viable options for the localization loss and allow a seamless integration of HBB representations to the propsed GBB-based losses.

\textbf{Licenses of the assets (code and datasets) used in our work}

    \resizebox{\textwidth}{!}{\begin{tabular}{c|c|c}
        \\
        asset &  license & Source \\ \hline
        EfficientDet~\cite{tan2020efficientdet} & Apache License 2.0 & \url{https://github.com/xuannianz/EfficientDet}\\
        SSD~\cite{Liu:ECCV:2016} & GNU Public License v3.0 & \url{https://github.com/Zzh-tju/DIoU-SSD-pytorch}\\
        UranusDet~\cite{yang2021rethinking} & Apache License 2.0 & \url{https://github.com/yangxue0827/RotationDetection}\\
        DOTA dataset~\cite{Xia:2018:CVPR}& academic purposes only & \url{https://captain-whu.github.io/DOTA/dataset.html}\\
        PASCAL VOC 2008~\cite{VOC2007}& academic purposes only & \url{http://host.robots.ox.ac.uk/pascal/VOC/} \\ 
        HRSC2016~\cite{liu:icpram:2017} & scientific research & \url{https://www.kaggle.com/guofeng/hrsc2016}
    \end{tabular}}

\bibliographystyle{authordate1}
\bibliography{references}

\end{document}